\ifdefined\pdfoutput
  \pdfoutput=1
\fi
\documentclass{article}

\usepackage{iclr2026_conference,times}
\iclrfinalcopy
\usepackage{xpeng-defs}
\usepackage{xpeng}  %
\setcitestyle{numbers,square,citesep={,}}
\usepackage{tikz}
\usepackage{wrapfig}
\usepackage{needspace}
\usepackage{placeins}
\usetikzlibrary{positioning}
\usepackage{hyperref}
\hypersetup{
  colorlinks=true,
  linkcolor=xpengblue,
  citecolor=xpengblue,
  urlcolor=xpengblue,
  filecolor=xpengblue,
  pdfborder={0 0 0},
}
\usepackage{url}
\usepackage{amsmath,amssymb}
\usepackage{empheq}
\usepackage{algorithm}
\usepackage{algorithmic}
\usepackage{fontawesome5}

\title{\modelname: Joint World and Action Modeling from Heterogeneous Experience}

\author{
\vspace{-15pt}\\
  \textbf{World Model Team, XPENG Robotics} \\
  \faGithub\enspace Project: \href{https://xpeng-robotics.github.io/xpace/}{\texttt{https://xpeng-robotics.github.io/xpace/}}
}

\setxpenglogo{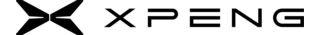}
\setxpenglab{XPENG Robotics}
\setxpengreport{}

\begin{document}
\raggedbottom

\maketitle

\begin{abstract}
A general-purpose robot needs to draw on diverse experience, choose actions, and anticipate how those actions will change the world.
We introduce \textbf{\modelname}, a unified embodied world model that serves as both a \emph{world action model}, jointly predicting executable robot actions and future video, and a \emph{world simulator}, predicting the visual consequences of prescribed actions.
Our key insight is that video prediction can both connect heterogeneous experience to action learning and generate new experience for policy improvement.
With a shared video backbone between the policy and simulator, we use action-unlabeled video to learn visual dynamics and action-labeled human and robot demonstrations to jointly learn video and action prediction.
Building on this architecture, a coarse-to-fine training curriculum progressively emphasizes robot control while retaining human experience, allowing the policy to learn behaviors beyond those covered by robot demonstrations.
Beyond learning from recorded experience, \textbf{\modelname} uses its simulator to create additional recovery supervision for the policy.
Specifically, we adapt the simulator to its own generated context, synthesize deviation--recovery trajectories around expert demonstrations, and fine-tune the policy on filtered recovery examples.
Experiments on XPENG's IRON humanoid robot show that heterogeneous training improves robustness and enables transfer of human-observed skills to tasks absent from robot demonstrations, while recovery data generated by the model's own simulator further improves real-world task completion.
Together, these results demonstrate how joint world and action modeling connects learning from heterogeneous experience with simulation-driven policy self-improvement.

\end{abstract}

\noindent\begin{minipage}{\linewidth}
  \centering
  \captionsetup{type=figure,font=small,skip=5pt}
  \includegraphics[width=\linewidth]{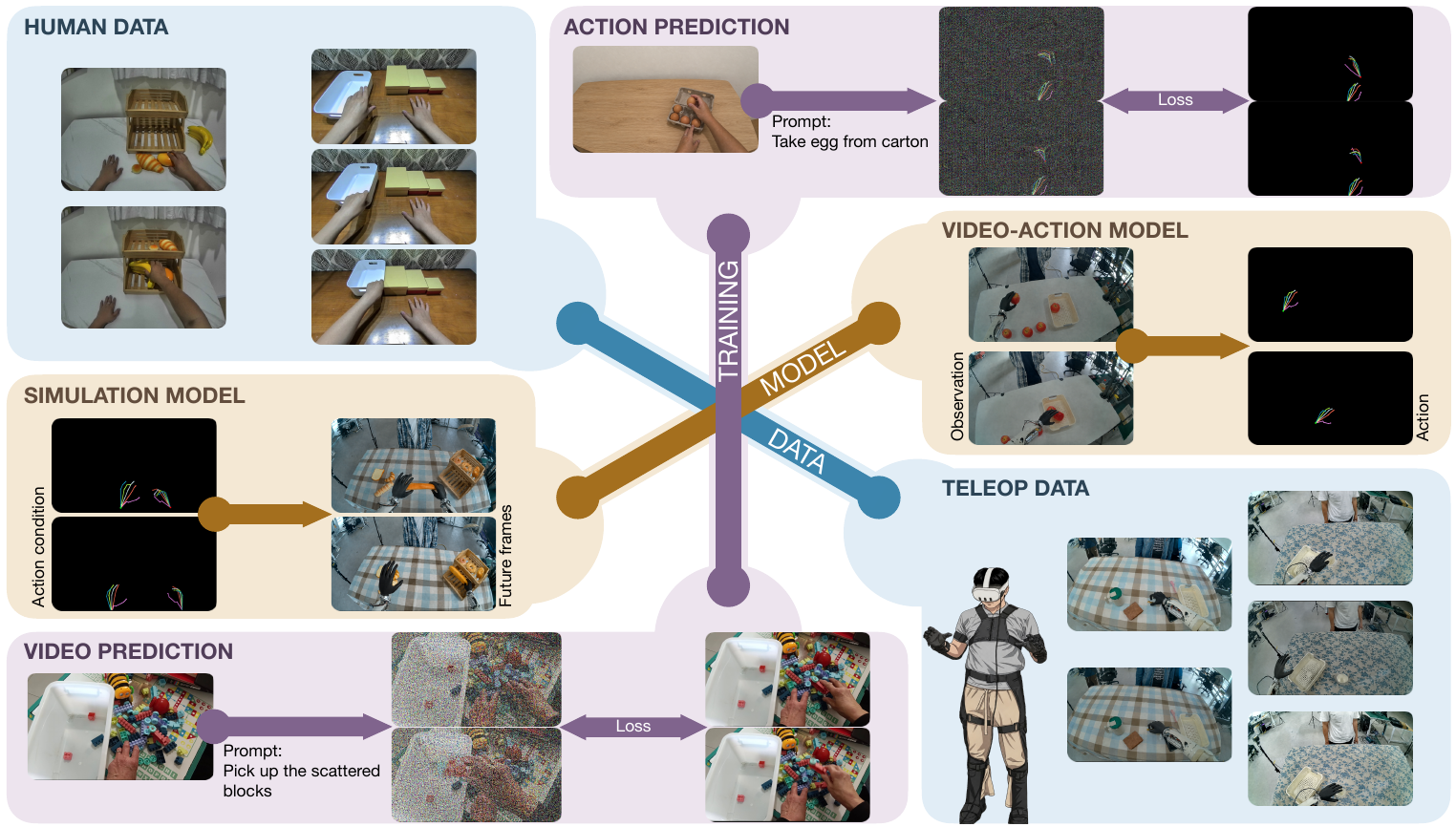}
  \caption{
  \modelname is a unified embodied world model, serving as both a \emph{world action model}, jointly predicting executable actions and future video from observations, current state, and task instructions, and a \emph{world simulator}, predicting future video from history, prescribed skeleton controls, and camera poses.
  The model learns jointly from human video and robot demonstrations, with
  human bridge data supporting cross-embodiment alignment.
  Recovery trajectories generated by the learned world simulator enable policy self-improvement through DAgger fine-tuning.}
  \label{fig:teaser}
  \vspace{-5pt}
\end{minipage}\par

\section{Introduction}
\label{sec:intro}

Humans acquire skills by observing others and interacting with the world, and use this experience to anticipate the consequences of their actions and refine their behavior.
A general-purpose embodied system should similarly learn from diverse experience, predict and execute actions, and improve its behavior beyond the demonstrations it has observed.
Progress in world modeling has brought prediction closer to robot control, from learned environment dynamics for planning and policy learning~\citep{ha2018worldmodels,wu2023daydreamer} to visually rich simulation of embodied interactions~\citep{yang2024unisim,zhu2024irasim}.
More recently, video--action models have connected future prediction and action generation within shared networks~\citep{zhu2025uwm,li2025uva}.
These developments provide important components of a general-purpose embodied system, but joint prediction alone does not establish how to acquire skills from diverse sources and improve them through simulated experience.

Closing this gap requires more than expanding a model's prediction outputs.
Human video offers broad behavioral coverage, but its supervision ranges from vision alone to motion annotations that differ from robot control commands; robot demonstrations provide executable actions but are costly to collect across diverse environments~\citep{mendonca2023swim,khazatsky2024droid}.
Simply combining these sources does not explain how their different supervision should contribute to a common policy and simulator.
Even with a shared model, training must adapt broad human experience to the target robot without discarding its behavioral coverage.
Furthermore, imitation on recorded demonstrations provides limited corrective supervision when execution errors lead the policy into unfamiliar states~\citep{ross2011dagger}; predictive simulation is useful for self-improvement only if its outputs can become reliable policy-training examples.
These considerations lead to three connected questions: \emph{(Q1) what experience should the model learn from; (Q2) how can one model use that experience for both action generation and simulation; and (Q3) how should training turn these capabilities into robot control and subsequent self-improvement?}

We introduce \textbf{\modelname}, a joint world--action framework that addresses these questions through coordinated data, model, and training designs, as illustrated in Figure~\ref{fig:teaser}.
It learns from heterogeneous human and robot experience and uses a shared video backbone to support both task-conditioned action generation and action-conditioned simulation.
The learned simulator then generates recovery supervision for policy fine-tuning, connecting learning from recorded experience with simulation-driven self-improvement within the same model.

\emph{Learning from complementary experience.}
To address Q1, we combine action-unlabeled egocentric video, kinematically labeled human demonstrations, and robot teleoperation rather than restricting policy learning to robot-collected experience.
Action-unlabeled videos supply diverse visual interactions, human motion labels supervise manipulation behaviors, and robot actions provide direct supervision for control on the target embodiment.
Task- and appearance-aligned human bridge data further reduce differences in task coverage and visual conditions between human demonstrations and robot deployment.
The resulting data pyramid connects broad interaction experience with robot-specific supervision while preserving the distinct information each source provides.

\emph{Joint action generation and simulation.}
To address Q2, we use video prediction as a common learning task across sources and connect it to action prediction through a shared video backbone.
In policy mode, observations, the current state, and task instructions condition the prediction of executable actions and future video; in simulation mode, visual history, prescribed skeleton controls, and camera poses condition future-video prediction.
An action Transformer reads the backbone's multi-level visual features, while action-loss gradients update the same backbone used by the simulator.
Embodiment-specific action projections accommodate human and robot action spaces while keeping the action Transformer shared.
Thus, video-only experience and action-labeled demonstrations can train the same model, with shared visual dynamics supporting both control and simulation.

\emph{Progressive training and simulation-driven self-improvement.}
To address Q3, we first train video prediction and then jointly train action prediction and simulation, applying action losses only where valid supervision is available.
A coarse-to-fine curriculum progressively emphasizes bridge and robot data while retaining human experience throughout robot adaptation.
Training then extends beyond the recorded demonstrations: we adapt the simulator to self-generated visual context through self-gradient forcing and use it to render prescribed deviation--recovery trajectories around expert demonstrations.
We filter the generated trajectories for recovery consistency, combine the retained recoveries with expert continuations, and fine-tune a separate policy copy on the augmented data.
This procedure uses the model's own simulator to supply corrective experience that is missing from nominal demonstrations, turning its predictive capability into a practical source of policy self-improvement.

Together, these designs contribute a framework that combines heterogeneous supervision, shared action and world modeling, and simulation-generated recovery training.
Experiments on XPENG's IRON humanoid robot show improved policy performance and robustness on the evaluated tasks, including transfer of human-observed skills to tasks absent from robot teleoperation.
Video-prediction experiments show that simulator adaptation improves autoregressive prediction while reducing inference cost, supporting recovery-data synthesis.
Fine-tuning on the synthetic recoveries further improves real-world task completion, demonstrating that the learned simulator can contribute to improving the model's own policy rather than serving only as a prediction interface.

\section{Related Work}
\label{sec:related_work}

\subsection{Data: human experience and robot grounding}

Human and robot datasets provide complementary supervision for embodied learning.
Ego4D captures diverse everyday activities from a human perspective \cite{grauman2022ego4d}, while Open X-Embodiment pools robot trajectories across embodiments \cite{openx2024} and DROID broadens robot scene and task coverage \cite{khazatsky2024droid}.
Human videos provide visual interaction experience, and tracked human demonstrations additionally supply motion labels; robot trajectories pair observations with executable controls.
Combining these sources requires accounting for differences in action representations, camera viewpoints, and task coverage.

Recent work studies how to transfer this human experience to robot control.
SWIM learns manipulation affordances from human videos before grounding them in robot interaction \cite{mendonca2023swim}, while Phantom and Human2Robot explore transferring human demonstrations to robot manipulation \cite{lepert2025phantom,xie2026human2robot}.
EgoScale combines human pretraining with aligned human--robot adaptation \cite{zheng2026egoscale}.
EgoWAM compares pixel, feature, and motion prediction for transfer under controlled human--robot co-training \cite{li2026egowam}.
Human experience can also train the predictive component: DreamDojo uses continuous latent actions to transfer interaction knowledge from large-scale egocentric video to robot world modeling \cite{gao2026dreamdojo}.
In this setting, \modelname{} jointly learns from action-unlabeled video, kinematically labeled human demonstrations, and robot trajectories, using task- and appearance-aligned bridge data to connect human experience with robot deployment.

Generated experience offers a complementary route to expanding robot data.
DreamGen synthesizes robot videos and recovers pseudo-actions with latent-action or inverse-dynamics models \cite{jang2025dreamgen}; RoboVIP augments manipulation observations through temporally coherent, multi-view generation guided by visual identity prompts \cite{wang2026robovip}.
These approaches expand behavioral or visual coverage, whereas our recovery augmentation targets off-trajectory states around recorded expert demonstrations.

\subsection{Model: policies, simulators, and shared predictive representations}

Robot foundation models connect perception and control through direct action prediction, visual simulation, or video-based planning.
RT-1 studies scalable transformer policies, while Octo learns a generalist policy that can adapt across robot observation and action spaces \cite{brohan2022rt1,octo2024}.
Diffusion Policy models multimodal action sequences through conditional denoising \cite{chi2023diffusionpolicy}.
Vision--language--action policies such as OpenVLA, $\pi_0$, and Gemini Robotics further connect pretrained visual--language understanding to executable control \cite{kim2024openvla,black2024pi0,geminirobotics2025}.
Their direct policy interface contrasts with approaches that explicitly expose predicted visual outcomes.

World simulators predict visual outcomes under supplied controls.
UniSim learns an interactive simulator from heterogeneous experience \cite{yang2024unisim}, and IRASim studies action-conditioned robot video generation \cite{zhu2024irasim}.
Ctrl-World combines frame-level action conditioning, multi-view prediction, and memory retrieval for extended policy interactions \cite{guo2025ctrlworld}.
AnyWorld factorizes action, camera, and embodiment conditions for cross-embodiment generation \cite{chen2026anyworld}; SyncWorld instead uses visual calibration episodes to interpret numerical actions in unseen setups without downstream training \cite{yang2026syncworld}.
These designs address complementary aspects of controllability and transfer across visual and embodiment changes.

Visual planning uses predicted observations as intermediate goals for control.
UniPi generates language-conditioned video plans and translates them into actions through an inverse-dynamics model \cite{du2023unipi}, while SuSIE uses image-editing diffusion to propose subgoals for a goal-conditioned controller \cite{black2023susie}.
RoboDreamer factorizes video generation for compositional instructions, and ManipDreamer introduces instruction trees and visual guidance \cite{zhou2024robodreamer,li2025manipdreamer}.
Video Prediction Policy extracts predictive features from a video diffusion model \cite{hu2024vpp}, whereas Video Policy combines video and action generation in an end-to-end trainable modular framework \cite{liang2025videopolicy}.

Unified video--action models connect prediction and control within shared networks.
UWM uses independent video and action noise levels for policy, dynamics, and video-generation modes \cite{zhu2025uwm}; UVA uses shared video--action latents with lightweight decoders and optional video decoding at deployment \cite{li2025uva}.
VideoVLA jointly forecasts actions and visual outcomes from language and images \cite{shen2025videovla}, while DreamZero transfers experience from video-only data through joint video--action modeling \cite{dreamzero2026}.
Motus combines understanding, video, and action experts \cite{motus2026}; Genie Envisioner builds policy and simulator modules around a pretrained video model \cite{liao2026genie}; and $\tau_0$-WM combines video--action prediction, action-conditioned simulation, and candidate evaluation \cite{zhou2026tau0wm}.
LingBot-VA couples autoregressive video--action modeling with asynchronous execution \cite{li2026lingbotva}, and Riemann-1.0 unifies policy execution and world simulation in a causal autoregressive sequence \cite{sun2026riemann}.
Cosmos Policy encodes actions, future images, and values as latent frames within a pretrained video model \cite{kim2026cosmospolicy}.

Other designs change the representation of actions or futures.
X-WAM adds spatial prediction and asynchronous video--action denoising \cite{guo2026xwam}; Action Images represents controls as pixel-grounded multiview videos \cite{zhen2026actionimages}; and FlowWAM uses optical flow as a shared motion representation for policy and world-model modes \cite{chen2026flowwam}.
Being-H0.7 learns future-informed latent queries without generating future frames at deployment \cite{beingh07_2026}.
Fast-WAM retains video co-training but skips test-time future prediction \cite{yuan2026fastwam}, while GigaWorld-Policy uses an action-centered causal design with optional video generation \cite{ye2026gigaworldpolicy}.
Together, these works distinguish the representational benefits of future prediction from the cost of explicitly generating futures during control.
Within this design space, \modelname{} shares a causal video backbone between policy and simulation, with an asymmetric action Transformer reading multi-level visual features and skeleton controls and camera poses conditioning simulation.

\subsection{Training: self-supervised video learning and supervised action grounding}

R3M learns manipulation representations through temporal and video--language objectives \cite{nair2023r3m}, while masked visual pretraining learns from egocentric images without action labels \cite{radosavovic2023mvp}.
Voltron combines language-conditioned visual reconstruction and visually grounded language generation to learn from human videos and captions \cite{karamcheti2023voltron}.
V-JEPA 2 follows large-scale video learning with robot-based action-conditioned prediction and planning \cite{assran2025vjepa2}.
LAPA learns discrete latent actions from frame transitions before adapting with action labels \cite{ye2025lapa}.
AdaWorld similarly extracts latent actions for world-model pretraining and subsequent adaptation to new controls \cite{gao2025adaworld}.
These objectives make action-unlabeled video useful without assuming that it already provides executable robot commands.
Our recipe combines video-only pretraining with joint video and action learning, retaining both human and robot action supervision as training increasingly emphasizes the target robot.

Autoregressive simulation additionally requires handling the shift from recorded to generated visual context.
Self Forcing trains video diffusion models on generated context \cite{huang2025selfforcing}, while Self Gradient Forcing updates context representations using future-prediction gradients without backpropagating through the full rollout \cite{zhuang2026selfgradientforcing}.
Persistent Robot World Models addresses multi-step drift through reinforcement-learning post-training on autoregressive rollouts \cite{bardhan2026persistent}.
Our simulator adaptation follows these principles while retaining paired, skeleton- and camera-conditioned flow-matching supervision to learn the visual consequences of prescribed controls (\S\ref{sec:stage-iii-sim}).

\subsection{Simulation-based policy improvement and evaluation}

Learning from imagined experience predates large video models.
Dreamer optimizes behavior through latent imagination \cite{hafner2019dreamer}, and model-based policy optimization uses short model rollouts branched from real data to balance additional experience against model error \cite{janner2019mbpo}.
Video world models extend this idea to visual manipulation, with different sources of supervision and different ways of updating policies.

DAgger aggregates expert labels on learner-visited states to address imitation-learning distribution shift \cite{ross2011dagger}.
WM-DAgger synthesizes and filters recovery trajectories around demonstrations \cite{yu2026wmdagger}, while Hi-WM collects human corrections during simulated policy rollouts \cite{li2026hiwm}.
RISE and World-VLA-Loop also use learned simulation for policy refinement \cite{yang2026rise,liu2026worldvlaloop}.
VLAW iteratively improves a world model with real policy rollouts and uses synthetic rollouts to improve the policy \cite{guo2026vlaw}.
PlayWorld collects autonomous play interactions to train a simulator for evaluation and model-based reinforcement learning \cite{yin2026playworld}.
More recently, Motus2 exposes policy, simulator, and value interfaces through shared weights and couples them for planning and model-based policy improvement \cite{bi2026motus2}.
Our focus is to obtain this corrective supervision from a simulator jointly trained with the policy.
After self-gradient forcing adaptation, \modelname{} renders prescribed deviation--recovery trajectories and fine-tunes a separate policy on filtered recoveries and expert data, following DAgger-style recovery augmentation rather than on-policy expert querying.

For evaluation, WorldGym tests whether simulated rollouts predict real-robot policy performance \cite{quevedo2025wpe}.
Our policy-conditioned rollouts support qualitative diagnosis of instruction following and target selection before testing; quantitative evidence for policy self-improvement comes from real-robot trials (\S\ref{sec:exp-policy}).
\section{Method}
\label{sec:method}

We present \textbf{\modelname}, a unified world and action modeling framework that integrates data, architecture, and training to learn robot control from heterogeneous experience and enable simulation-driven policy self-improvement.
To combine the breadth of human experience with the precision of robot demonstrations, we organize these sources into a four-layer data pyramid (\S\ref{sec:data}).
Learning from this mixture requires an architecture that connects visual prediction with executable actions; we therefore share a video backbone between the simulator and the policy, allowing both to learn from a common representation of visual dynamics (\S\ref{sec:model}).
Within this architecture, a coarse-to-fine curriculum progressively emphasizes robot supervision while retaining human experience, adapting the shared representation to the deployment embodiment (\S\ref{sec:training}).
The resulting model supports learning beyond recorded demonstrations: after self-gradient-forcing adaptation, its simulator generates recovery trajectories that provide additional supervision for policy fine-tuning, completing the simulation-driven self-improvement process (\S\ref{sec:post-training}).

\subsection{Data}
\label{sec:data}
Our data design combines human and robot experience with language conditions and action representations that support learning across sources and embodiments.
We organize $5{,}000$ hours of embodied video into four layers: (L1) broad egocentric video for learning visual dynamics, (L2) human video--action data for learning manipulation behaviors, (L3) bridge data for aligning human and robot experience, and (L4) robot teleoperation for learning embodiment-specific control.
A separate failure set (F) supplies unsuccessful, low-progress, and recovery trajectories exclusively for simulator training.
To associate the recorded behaviors with task instructions, we construct multi-level captions that describe both episode-level objectives and clip-level actions, with caption sampling progressively adapted to policy deployment.
We further express human and humanoid robot motion in a shared action schema, allowing demonstrations from both embodiments to supervise action prediction while preserving embodiment-specific mappings.

\begin{figure}[t]
  \centering
  \includegraphics[width=\linewidth]{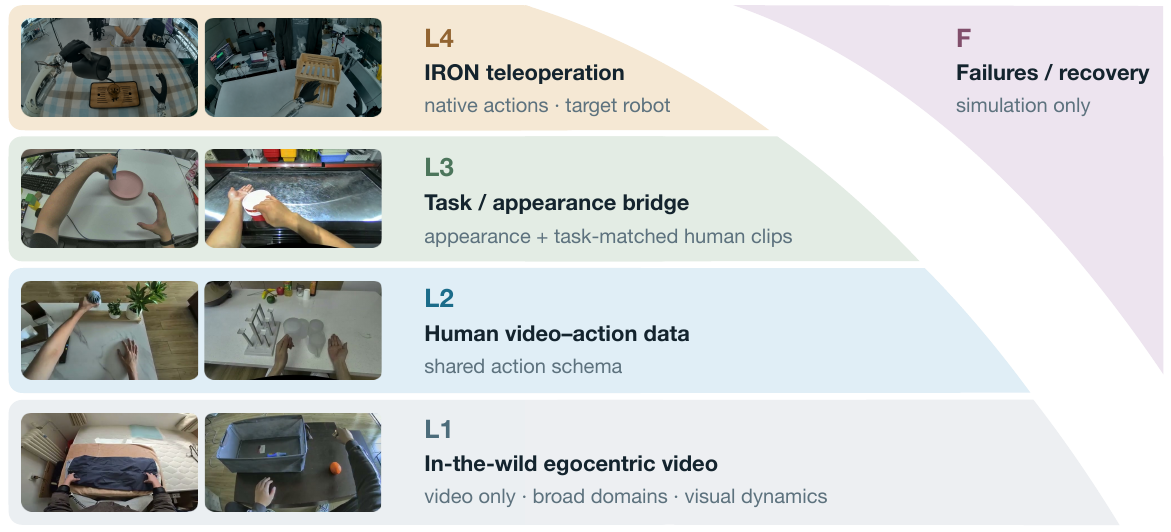}
  \caption{\textbf{Data pyramid for \modelname.}
  The corpus comprises (L1) in-the-wild egocentric video, (L2) human video--action data, (L3) task- and appearance-aligned bridge data, and (L4) IRON teleoperation.
  Failure and recovery data (F) train simulation only.}
  \label{fig:data_pyramid}
\end{figure}

\paragraph{Data pyramid.}
As shown in Figure~\ref{fig:data_pyramid}, the data pyramid connects broad interaction coverage with robot-executable control through four complementary layers:
\begin{enumerate}
  \renewcommand{\labelenumi}{\textbf{(L\arabic{enumi})}}
  \item \textbf{In-the-wild egocentric video.}
  We select interaction-rich, manipulation-relevant clips to provide broad coverage of scenes, objects, and visual dynamics.
  These clips have no action labels and train the model to predict future video.

  \item \textbf{Human video--action data.}
  Egocentric videos paired with hand and wrist poses provide action supervision across diverse manipulation behaviors.
  The poses use the shared human--robot action schema, enabling the model to learn action prediction from human demonstrations alongside future-video prediction.

  \item \textbf{Task- and appearance-aligned bridge data.}
  We construct bridge data to reduce task and visual differences between human demonstrations and robot deployment.
  \emph{Task-aligned} clips are selected from L1 to match manipulation task families in the robot corpus, concentrating human experience on robot-relevant skills.
  \emph{Appearance-aligned} clips are collected in environments resembling the robot workspace in camera viewpoint, background, and object arrangement, without requiring exact task matching.
  Joint training on these sources encourages cross-embodiment representations through task and visual overlap, without an explicit feature-alignment loss.

  \item \textbf{Robot teleoperation.}
  Demonstrations collected on IRON robots, including the deployment embodiment, pair video with native robot actions.
  They provide direct supervision for executing manipulation tasks on the target hardware.
\end{enumerate}

\paragraph{Failure and recovery data.}
Training a simulator exclusively on successful demonstrations can bias its predictions toward successful outcomes even when the supplied actions are incorrect.
We therefore include failure, low-progress, and recovery trajectories in F to supervise the visual consequences of both successful and unsuccessful motions.
These trajectories train video prediction conditioned on recorded skeleton controls and camera poses; their actions are excluded from policy imitation.
This supervision supports action-dependent simulation for policy evaluation and recovery-data synthesis.

\begin{figure}[t]
  \centering
  \begin{minipage}[t]{0.575\linewidth}
    \centering
    \vspace{0pt}
    \includegraphics[width=\linewidth]{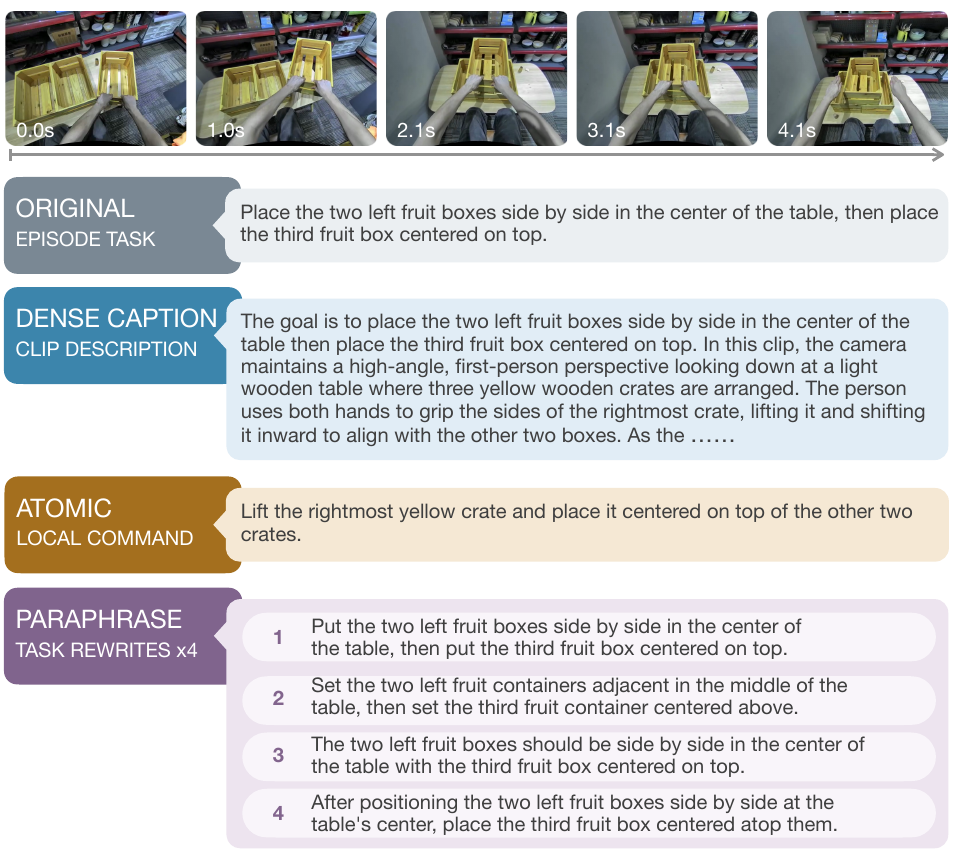}
    \par\smallskip
    (a)
  \end{minipage}\hfill
  \begin{minipage}[t]{0.415\linewidth}
    \centering
    \vspace{0pt}
    \includegraphics[width=\linewidth]{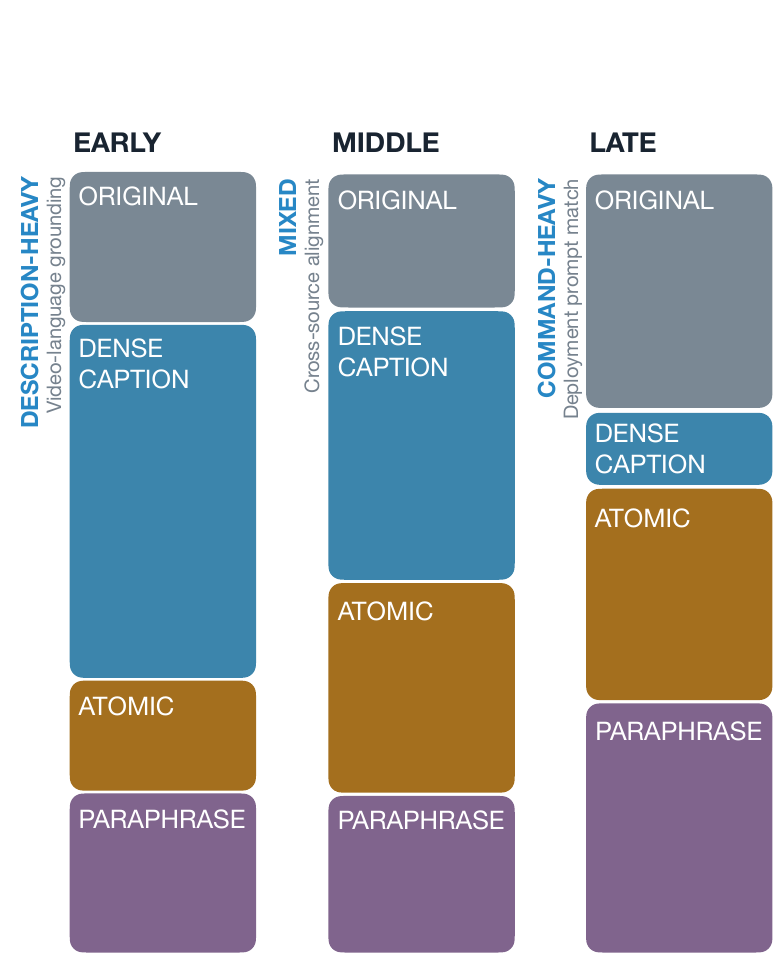}
    \par\smallskip
    (b)
  \end{minipage}
  \caption{\textbf{Multi-level language conditioning and caption sampling.}
  (a)~Each training clip is associated with an original task instruction, a dense description, an atomic action command, and task paraphrases, covering the overall objective, local actions, and alternative wording.
  (b)~Training samples one text condition per example, with the mixture shifting from dense descriptions toward task instructions to match deployment inputs.}
  \label{fig:caption_levels}
\end{figure}

\paragraph{Multi-level captions.}
Language provides task instructions for the policy and should accurately describe the actions demonstrated in each training clip.
Each training clip is a short temporal segment sampled from a full task episode and may contain only part of the instructed task.
We therefore retain the episode-level instruction and supplement it with clip-specific captions and task paraphrases, as shown in Figure~\ref{fig:caption_levels}:
\begin{enumerate}
  \item \textbf{Original task instruction.}
  The source instruction specifies the overall task objective, preserving the context of the episode.

  \item \textbf{Dense caption.}
  A detailed description specifies the scene and action sequence within the sampled clip, using the descriptive language typical of video-model pretraining.

  \item \textbf{Atomic action caption.}
  A short imperative command specifies the local action demonstrated in the clip, providing precise action guidance when the original instruction describes a longer task.

  \item \textbf{Task paraphrases.}
  LLM-generated rewrites preserve the original task objective while varying its wording, exposing the model to multiple expressions of the same instruction.
\end{enumerate}

These text conditions differ in both temporal scope and linguistic style, allowing training to connect detailed video descriptions with concise task instructions.
We sample one condition per example using a stage-dependent mixture (\S\ref{sec:training}).
Early training emphasizes dense captions to remain close to the pretrained video model's text-conditioning distribution; subsequent stages progressively increase original and paraphrased task instructions to match policy deployment.
L1 clips without action labels retain descriptive captions throughout training for language-conditioned video prediction.

\paragraph{Cross-embodiment action schema.}
Our humanoid robot provides a direct correspondence between human and robot arm and hand motion, facilitating cross-embodiment action alignment.
Unlike approaches that transfer human demonstrations to gripper-based robots~\citep{lepert2025phantom,xie2026human2robot}, our shared action representation retains hand articulation alongside wrist motion.
We therefore represent human and robot trajectories using a common kinematic schema comprising end-effector poses, hand articulation, and torso/camera poses, with source-specific normalization.
Human and robot demonstrations supervise the same action-prediction objective $\mathcal{L}_a$, enabling human motion data to contribute directly to policy learning.
To accommodate differences in embodiment kinematics and action dimensions, we use embodiment-specific linear input and output projections with a shared action Transformer, as described in \S\ref{sec:model}.
This design combines shared action modeling with embodiment-specific mappings for executable control.

\subsection{Model Architecture}
\label{sec:formulation}
\label{sec:model}

\begin{figure}[t]
  \centering
  \includegraphics[width=\linewidth]{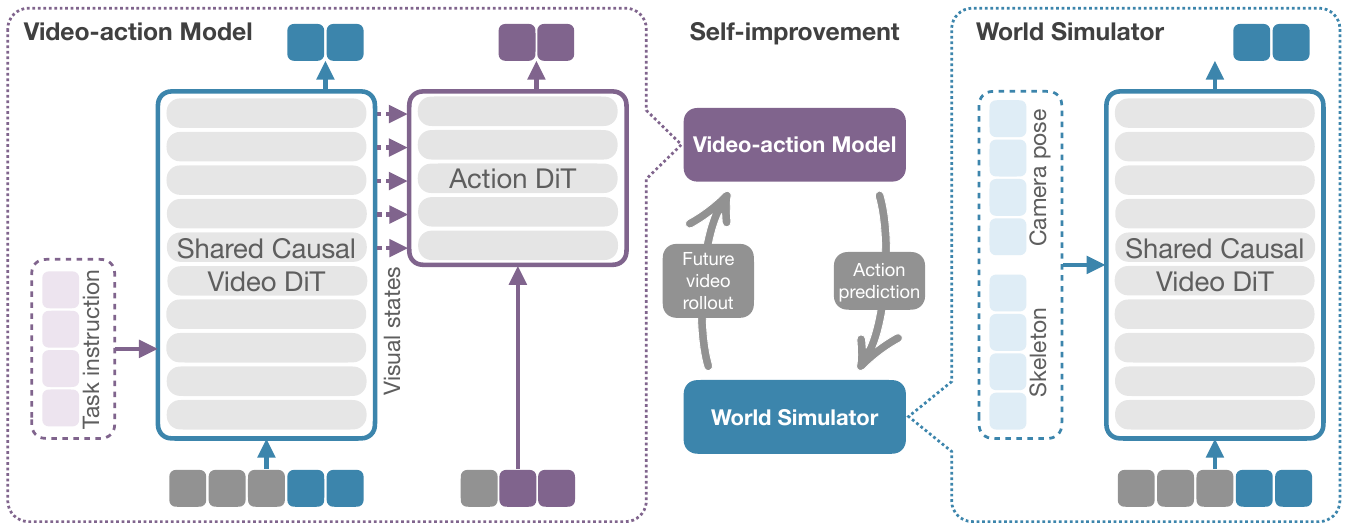}
  \caption{\textbf{MoT architecture, operating modes, and self-improvement of \modelname.}
  The asymmetric mixture-of-transformers (MoT) couples a causal video
  Transformer with an action Transformer through multi-level video-to-action
  feature connections.
  \textbf{Left:} In video--action mode, visual history, the current state, and a task instruction condition the joint prediction of future video and executable action chunks.
  \textbf{Right:} In simulation mode, the shared video Transformer predicts future video from visual history, prescribed skeleton controls, and camera poses.
  \textbf{Center:} Simulation connects action sequences to their visual
  consequences, enabling simulation-driven policy self-improvement.
  After adaptation with self-gradient forcing, the simulator generates
deviation--recovery trajectories that are filtered and aggregated with
training data for DAgger-style policy fine-tuning.}
  \label{fig:model_arch}
\end{figure}

The heterogeneous data in \S\ref{sec:data} provide both visual dynamics and cross-embodiment action supervision; our architecture connects these signals through a shared predictive representation.
As in Figure \ref{fig:model_arch}, \modelname uses an asymmetric mixture-of-transformers (MoT) architecture: a Video Transformer predicts future video, and an Action Transformer reads its multi-level visual features to predict executable actions.
The video backbone and its output head are shared between policy and simulation modes, while embodiment-specific action projections accommodate the human and robot action spaces.
Information flows from video to action in the forward pass, but action-loss gradients propagate into the video backbone during joint training, allowing control supervision to refine the representation used by both modes.

\paragraph{Chunk-wise prediction.}
Both the video action model and the world simulator are trained with the same chunk-wise video representation and causal attention structure.
A frozen video VAE encodes each clip into an initial anchor latent $z_0$ followed by $M$ chunks $\mathbf{z}^{1:M}$, each containing 4 latent frames (corresponding to 16 video frames).
For prediction chunk $m$, the visual context consists of $z_0$ and the preceding chunks $\mathbf{z}^{<m}$; this prefix is empty when $m=1$.
We use \emph{chunk-wise causal attention}: attention is bidirectional within each chunk, but access across chunks is restricted to preceding context.
The corresponding policy action chunk is $\mathbf{a}^m\in\mathbb{R}^{N_a\times D_a}$, where $N_a=16$ is the action horizon and $D_a$ is the embodiment-specific action dimension; $\kappa^m$ denotes the aligned 2D action-skeleton sequence.
We denote the current state at the start of chunk $m$ by $\mathbf{s}^m$ and the camera-pose sequence aligned with the chunk by $\mathbf{c}_{\mathrm{cam}}^m$.
During teacher-forced training, the visual context comes from recorded  ${\mathbf{z}}^{<m}$; SGF replaces the recorded prefix with self-generated chunks $\widehat{\mathbf{z}}^{<m}$, as described in \S\ref{sec:stage-iii-sim}.

\paragraph{Two operating modes.}
Under this shared chunk-wise formulation, the two modes differ in their conditions and active modules:
\begin{itemize}
  \item \textbf{World simulator.}
  Given visual history, a prescribed 2D action-skeleton sequence $\kappa^m$, and camera poses $\mathbf{c}_{\mathrm{cam}}^m$, the video Transformer predicts the visual consequences of the supplied motion:
  \begin{equation}
  p(\mathbf{z}^m\mid z_0,\mathbf{z}^{<m},\kappa^m,\mathbf{c}_{\mathrm{cam}}^m).
  \end{equation}
  The action Transformer is inactive, and no language instruction is supplied.

  \item \textbf{Video action model.}
  Given visual history and a language instruction $g$, the video Transformer predicts a task-conditioned future and supplies its hidden features to the action Transformer, which additionally receives the current state $\mathbf{s}^m$:
  \begin{equation}
  p(\mathbf{z}^m,\mathbf{a}^m\mid z_0,\mathbf{z}^{<m},\mathbf{s}^m,g,e),
  \end{equation}
  where $e$ selects the embodiment-specific action projections.
  Prescribed future skeleton and camera-pose controls are withheld, so the policy must infer motions from visual history, the current state, and the instruction.
\end{itemize}

\paragraph{Skeleton- and camera-conditioned simulation.}
The simulator receives camera poses as an explicit input alongside the skeleton sequence, rather than using them only to render the skeleton.
We incorporate skeleton controls through \emph{token addition} at the Transformer input.
The rasterized skeleton sequence is first encoded by the frozen video VAE, then mapped through a separate patch embedding to match the video tokens in spatial and temporal layout and embedding dimension.
Each skeleton token is added element-wise to the video token at the corresponding spatial and temporal position, producing a single conditioned token sequence without increasing its length.
The skeleton patch embedding is zero-initialized to preserve the pretrained video representation at initialization.
We compare token addition with other conditioning strategies in the ablation in \S\ref{sec:exp-actcond}.
During autoregressive simulation, generated chunks $\widehat{\mathbf{z}}^{<m}$ provide the visual context for predicting $\widehat{\mathbf{z}}^m$, while skeleton controls $\kappa^m$ and camera poses $\mathbf{c}_{\mathrm{cam}}^m$ remain externally supplied.

\paragraph{Asymmetric video-to-action policy model.}
The policy predicts actions from task-conditioned video features through three complementary designs.
A learned mode embedding replaces skeleton conditioning on the video tokens, while text cross-attention supplies the task instruction.
\begin{itemize}
  \item \textbf{Asymmetric multi-level action readout.}
  The 15 action blocks read spatially dense features from video blocks 16--30, respectively, to predict the action chunk $\mathbf{a}^m$.
  Each action block projects its paired video features into the action hidden dimension, then uses action tokens as queries and the concatenated action and projected video tokens as keys and values.
  This combines action self-attention with multi-level visual conditioning without spatial pooling.
  Action tokens do not modify video features in the forward pass, but action-loss gradients propagate through the visual readout into the video backbone.

  \item \textbf{Embodiment-indexed boundary projections.}
  Each embodiment has its own linear input projection and final output projection for actions.
  The action Transformer and all other action-branch components are shared across embodiments.

  \item \textbf{Current-state conditioning for the action Transformer.}
  The action Transformer prepends the normalized current robot state token $\mathbf{s}^m$ to the $N_a=16$ noisy action tokens and processes them with the existing input projection and shared Transformer.
  Discarding the state-token output leaves the 16-step action prediction without adding action-head parameters.
  During training, we randomly omit the current robot state condition for each sample with probability $0.5$.
\end{itemize}

\textbf{Real-robot deployment.}
We deploy \modelname{} for closed-loop control of the IRON robot, using streaming visual encoding and asynchronous action execution.`

\begin{itemize}

\item \textbf{Streaming VAE encoding} processes incoming observations incrementally, producing one latent frame per four new video frames while the robot continues executing queued commands.
After 4 video frames, the policy uses the accumulated visual latents, current robot state, and task instruction to generate a new 16-step action plan.
This separates the cadence of visual encoding from full policy inference, while observation sampling and command publication remain independent of the inference worker.

\item \textbf{Overlapping-horizon stitching} connects successive plans without pausing execution for inference.
Predicted actions are converted to robot control units and interpolated for command publication at approximately 10\,Hz.
The publisher preserves already committed commands, discards the overlapping prefix of a new plan according to the remaining queue length, and appends its non-overlapping suffix.
A decaying correction smooths the transition at the join, allowing new predictions to enter the running trajectory through publisher-side post-processing.
\end{itemize}

\subsection{Training Paradigm}
\label{sec:training}

\begin{figure}[t]
  \centering
  \includegraphics[width=\linewidth]{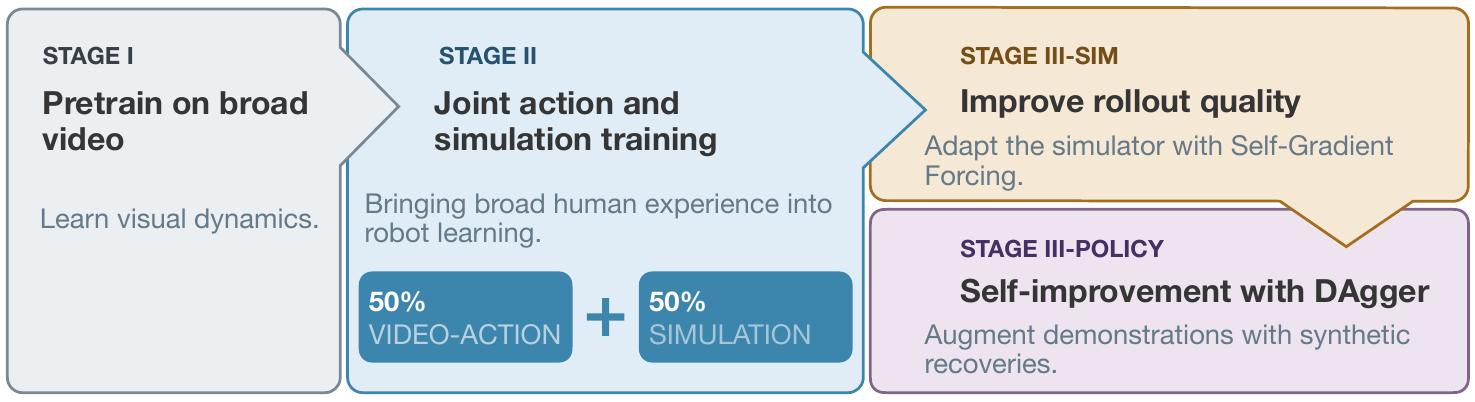}
  \caption{\textbf{Training stages and policy self-improvement of \modelname.}
  Stage~I trains video prediction; Stage~II adds the action branch and jointly trains policy and simulation modes with a coarse-to-fine data curriculum.
  Stage~III-sim and Stage~III-policy initialize separate copies of the Phase~II-c checkpoint, updating only simulation and policy, respectively.
  Stage~III-sim adapts the simulator to self-generated context, while the unchanged Phase~II-c policy serves as the Before DAgger baseline.
  The connection from Stage~III-sim to Stage~III-policy represents synthetic recovery data, not checkpoint initialization; the SGF simulator remains fixed during policy fine-tuning.}
  \label{fig:train_stages}
\end{figure}

Before post-training, we train the shared architecture in two stages, as shown in Figure~\ref{fig:train_stages}.
Stage~I adapts the video backbone to embodied video; Stage~II adds action learning and jointly trains policy and simulation modes, progressively increasing robot supervision while retaining human and video data.
The resulting model initializes the simulator adaptation and policy improvement described in \S\ref{sec:post-training}.

\paragraph{Training objectives.}
We train video and action prediction with flow-matching losses~\citep{lipman2023flowmatching}, denoted by $\mathcal{L}_v$ and $\mathcal{L}_a$, respectively.
Flow times and Gaussian noise are sampled independently for video and action.
Simulation updates optimize $\mathcal{L}_v$, while video--action updates optimize $\mathcal{L}_v+\mathcal{L}_a$ with equal loss weights.
Video loss is averaged over prediction chunks only; action loss is averaged over valid action-labeled samples and is zero when none are present.
Action-unlabeled clips contribute only to video prediction, and F is used exclusively for simulation.
Stages~I and~II use teacher forcing with recorded visual context.

\subsubsection{Stage~I: Video-only pretraining}
\label{sec:stage-i}

Starting from the pretrained video Transformer, we train future-video prediction on egocentric human video and robot teleoperation at near-natural source-mixture ratios.
All participating sources are treated as video-only data, with multi-level captions as text conditions and no action loss; F and DAgger data are excluded.
This stage lets the backbone learn from broad visual experience before action supervision is introduced.
We evaluate its benefit for subsequent policy learning in \S\ref{sec:exp-data}.

\subsubsection{Stage~II: Coarse-to-fine joint training}
\label{sec:stage-ii}

\begin{figure}[t]
  \centering
  \includegraphics[width=\linewidth]{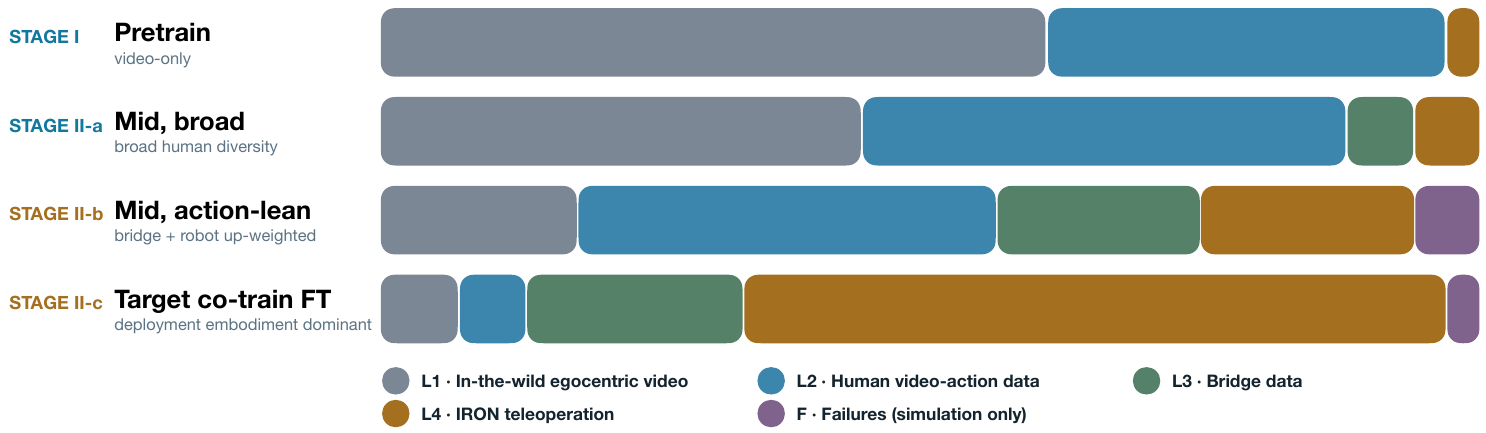}
  \caption{\textbf{Source-mixture curriculum through Stages~I and~II.}
  Segment widths indicate configured sampling weights; exact values are omitted.
  Stage~I uses the participating sources as video-only data.
  Across Stage~II, broad L1/L2 data are reduced while L3 bridge and L4 robot data gain weight, with target-robot data dominant in II-c.
  F enters in II-b through simulation updates only.
  These source proportions are distinct from the fixed 50/50 sampling of video--action and simulation modes.}
  \label{fig:data_mixture_curriculum}
\end{figure}

We load the Stage~I video weights and initialize the entire action branch, including the action Transformer, visual projections, and embodiment-specific input and output projections.
At each update, we sample video--action or simulation mode with equal probability and optimize the corresponding objective.
The mode-sampling probabilities and training objectives remain unchanged across all three phases, while the data mixture changes as shown in Figure~\ref{fig:data_mixture_curriculum}.
Each phase continues from the full preceding checkpoint without reinitializing any modules or embodiment-specific projections.

\begin{itemize}
  \item \textbf{Phase~II-a: Broad co-training.}
  Broad human and action-unlabeled video dominate the mixture, while all action-labeled human and robot sources contribute to action prediction.

  \item \textbf{Phase~II-b: Increased robot and bridge supervision.}
  We increase robot teleoperation and bridge data, retain a reduced video-only stream, and introduce F in simulation updates only.

  \item \textbf{Phase~II-c: Target-robot co-training.}
  Target-robot data dominate, with a small replay mixture of human demonstrations, bridge data, and action-unlabeled video retained rather than switching to robot-only fine-tuning.
\end{itemize}

Replay maintains human action supervision and video prediction during robot adaptation to limit overfitting and forgetting.
We evaluate this through matched robot-only and human-mixed continuations in \S\ref{sec:exp-sim-cotrain} and analyze transfer across task categories in \S\ref{sec:exp-ood-category}.
For language-conditioned video--action updates, caption sampling shifts from dense descriptions toward original and paraphrased task instructions by II-c, following Figure~\ref{fig:caption_levels}; L1 clips retain descriptive captions.
Simulation is therefore trained throughout Stage~II; Stage~III-sim adapts it to generated visual context rather than introducing it for the first time.

\subsection{Post Training}
\label{sec:post-training}

Post-training aims to improve the policy through world-model-generated
corrective supervision.  Although Stages~I--II establish action generation
and trajectory-conditioned simulation, policy behavior cloning primarily
covers successful expert demonstrations.  It therefore provides limited
recovery supervision for off-trajectory states encountered during
closed-loop execution.  We adopt a DAgger-style approach: synthesize
deviation--recovery trajectories around expert demonstrations using the
world model, then aggregate the recovery examples with expert data for
policy fine-tuning.

The effectiveness of this approach depends on the simulator faithfully
following prescribed motions under its own generated context.  However,
teacher-forced training uses clean recorded histories, whereas
autoregressive inference conditions on previous predictions, allowing
errors to accumulate.  We therefore first adapt the simulator with
self-gradient forcing in \textbf{Stage~III-sim}.  The adapted world model
then synthesizes and filters recovery trajectories for policy updating in
\textbf{Stage~III-policy}.  This ordered procedure links simulator
adaptation, DAgger data synthesis, and policy refinement in a
world-model-driven self-improvement loop.

\subsubsection{Stage~III-sim: Chunk-wise self-gradient forcing}
\label{sec:stage-iii-sim}

We adopt self-gradient forcing (SGF)
\citep{huang2025selfforcing,zhuang2026selfgradientforcing} to adapt the
simulator from teacher-forced training to self-generated context.
SGF initializes a separate copy of the Phase~II-c checkpoint and optimizes only the simulation objective, with the action Transformer inactive.
The Phase~II-c checkpoint is preserved for base-policy evaluation and subsequent policy initialization.
We retain the chunk-wise representation defined in \S\ref{sec:model}: an observed anchor $z_0$ followed by $M$ prediction chunks $\mathbf{z}^{1:M}$.
For chunk $m$, the simulator condition is $c^m=(\kappa^m,\mathbf{c}_{\mathrm{cam}}^m)$, comprising the aligned 2D action-skeleton sequence and camera poses; no language condition is used.
We combine self-conditioned rollout generation with context-aware flow matching against the corresponding recorded targets.

\paragraph{Self-conditioned rollout generation.}
Starting from $z_0$, we generate chunks autoregressively using an
eight-step ODE sampler $\Psi_\theta^K$:
\begin{equation}
\widehat{\mathbf{z}}^m
=
\Psi_\theta^K\!\left(
\epsilon_z^m\mid z_0,\widehat{\mathbf{z}}^{<m},c^m
\right),
\qquad
\epsilon_z^m\sim\mathcal{N}(0,I),
\qquad K=8.
\end{equation}
Here $\widehat{\mathbf{z}}^{<m}$ contains all previously generated chunks and is empty for the first prediction.
Only $z_0$ is taken from the recorded video; subsequent visual context is generated by the simulator, while skeleton controls and camera poses remain externally supplied.
The complete rollout is executed without gradient tracking, using the
same autoregressive context and sampling budget as SGF inference.

\paragraph{Context-aware flow-matching optimization.}
We detach the generated prefix and re-encode it with gradients so that training updates the context learner without backpropagating through the rollout sampler.
For ground-truth chunk $\mathbf{z}^m$, we form $\mathbf{z}^m_{u_z}=(1-u_z)\epsilon_z+u_z\mathbf{z}^m$, with $\epsilon_z\sim\mathcal{N}(0,I)$ and target velocity $v^{z,m\star}=\mathbf{z}^m-\epsilon_z$.
The objective is
\begin{equation}
\mathcal{L}_{\mathrm{III\text{-}sim}}
=
\frac{1}{M}
\sum_{m=1}^{M}
w_m\,
\mathbb{E}_{u_z,\epsilon_z}
\left[
\left\|
v^z_\theta
\left(
\mathbf{z}^m_{u_z},u_z,c^m
\mid
z_0,\operatorname{sg}(\widehat{\mathbf{z}}^{<m})
\right)
-v^{z,m\star}
\right\|_2^2
\right],
\end{equation}
where $w_m$ weights each chunk and $\operatorname{sg}$ stops gradients
through generated latent values.
Rather than adopting a distribution-matching objective such as
DMD~\citep{yin2024dmd}, we retain $\ell_2$ loss:  
trajectory-conditioned simulation must reproduce the consequences of the
supplied controls, not merely generate plausible continuations.
Distribution matching alone does not enforce this sample-level
correspondence.
This adaptation prepares the simulator
for the deviation--recovery synthesis used in Stage~III-policy.
Section~\ref{sec:exp-rollout} evaluates its effect on prediction fidelity
and inference cost over short- and long-horizon rollouts.

\subsubsection{Stage~III-policy: Improve from aggregated corrections}
\label{sec:stage-iii-policy}
Self-gradient forcing turns the simulation slice into a stable video predictor. We next use this improved world
model to expand the policy training distribution beyond the successful expert
trajectories observed in L4.  Following the DAgger principle and
world-model-based recovery synthesis~\citep{ross2011dagger,yu2026wmdagger}, we
construct off-trajectory states around expert demonstrations, synthesize their
visual consequences with the simulation slice, and retain only trajectories
that return consistently to the expert manifold.

\paragraph{Deviation--recovery trajectory synthesis.}
Given an expert trajectory of length $T$, we choose a pivot frame $t_p$ and identify the arm with the largest cumulative end-effector displacement as the arm to perturb.
We displace its end-effector position along a sampled unit direction $v_d\in\mathbb{R}^3$, using step magnitude $\mu$ and deviation horizon $k\in\{8,16\}$.
The augmented position at synthesized step $j$ is
\begin{equation}
p^{\mathrm{aug}}_{t_p+j}
=
p_{t_p}+\mu d(j)v_d,
\qquad
d(j)=
\begin{cases}
j, & 1\le j\le k,\\
2k-j, & k<j\le2k.
\end{cases}
\end{equation}
The first $k$ frames move the selected end effector away from the expert trajectory, reaching the deviation endpoint at $t_p+k$; the next $k$ frames return it to the pivot pose at $t_p+2k$.
We retain directions whose angle to the subsequent expert motion lies between $90^\circ$ and $150^\circ$, avoiding both negligible perturbations and directly contradictory recovery supervision.
The other arm, rotations, camera state, and remaining channels are held at their pivot values during the synthesized segment.

\paragraph{World-model rendering and expert-tail composition.}
We render a 2D skeleton sequence from the augmented state trajectory and use the real video up to the pivot as visual context.
Conditioned on this skeleton sequence and the corresponding camera poses, held at their pivot values, the Stage~III-sim checkpoint generates the visual deviation and recovery.
We use the world model only for this newly inserted $2k$-frame segment: the prefix before the pivot and the task continuation after recovery are taken from the original expert video.
The resulting clip has the source layout:
\begin{equation}
\underbrace{o_{0:t_p}}_{\mathrm{GT\ prefix}}
\;\Vert\;
\underbrace{\hat o_{t_p+1:t_p+2k}}_{\mathrm{WM\ deviation+recovery}}
\;\Vert\;
\underbrace{o_{t_p+1:T}}_{\mathrm{GT\ expert\ tail}}.
\end{equation}
This composition prevents long-horizon world-model drift from corrupting the task continuation while preserving a visually grounded transition from an off-trajectory state back to successful behavior.
The policy-training crop starts at the deviation endpoint $t_p+k$, so each example contains the difficult state, the complete recovery, and the subsequent expert task progress.

\paragraph{Consistency-guided filtering.}
World-model rollouts may contain weak perturbations, incomplete recovery, or visual discontinuities.
We therefore compare the deviation endpoint, recovery endpoint, and expert pivot using DINOv3 feature similarity, PSNR, and SSIM\footnote{We compute semantic similarity using a frozen DINOv3-Large visual encoder~\citep{simeoni2025dinov3}. For PSNR and SSIM, inverse-depth weighting emphasizes nearby robot--object interaction regions while down-weighting the background.}.
Let $\mathcal{Q}=\{q_{\mathrm{DINO}},q_{\mathrm{PSNR}},q_{\mathrm{SSIM}}\}$ denote the three quality metrics.
For every $q\in\mathcal{Q}$, an eligible rollout must satisfy:
\begin{subequations}
\label{eq:dagger_filter}
\begin{empheq}[left=\empheqlbrace]{align}
\Delta_q
&\triangleq
q(\hat o_{t_p+2k},o_{t_p})-q(\hat o_{t_p+k},o_{t_p})>0,
\label{eq:dagger_filter_relative}
\\
R_q
&\triangleq
q(\hat o_{t_p+2k},o_{t_p})\ge\tau_q.
\label{eq:dagger_filter_absolute}
\end{empheq}
\end{subequations}
Equation~\eqref{eq:dagger_filter_relative} is a relative recovery criterion: the recovery endpoint must be closer to the expert pivot than the deviation endpoint.
Equation~\eqref{eq:dagger_filter_absolute} is an absolute consistency criterion, where $\tau_q$ is the acceptance threshold for metric $q$; it ensures that the recovery endpoint is sufficiently similar to the expert pivot to support a smooth transition into the expert tail.
Among eligible candidates, we convert the DINOv3, depth-weighted PSNR, and depth-weighted SSIM margins into within-pool percentile ranks and average them into a composite quality score.
We retain the highest-scoring candidates and construct a task-balanced mixture across the $k=8$ and $k=16$ recovery horizons.
The resulting recovery-tail clips constitute the DAgger dataset $D_{\mathrm{dagger}}$.

\paragraph{DAgger aggregation and policy update.}
Stage~III-policy initializes a separate policy copy directly from Phase~II-c, not from the SGF-adapted checkpoint.
Only this policy copy is fine-tuned; the SGF simulator remains fixed.
We retain the Phase~II-c data recipe (\S\ref{sec:stage-ii}) and mix in
$D_{\mathrm{dagger}}$ for policy fine-tuning.  DAgger recovery examples
account for $8\%$ of the training mixture, with the remaining data drawn
from the Phase~II-c sources.  This retains the original robot supervision
and replay data while adding corrective examples.
Training on this mixture preserves nominal expert behavior while exposing the policy to off-distribution states paired with successful recovery trajectories, thereby improving policy robustness under closed-loop execution.

\section{Experiments}
\label{sec:experiments}

\subsection{Experimental Setup}
\label{sec:exp-setup}

\paragraph{Setup.}
We evaluate video prediction quality and real-robot policy performance, and examine the effects of training data and simulation-generated recovery supervision on policy learning.
The base \modelname checkpoint serves as both a video simulator and a robot policy; the self-improvement experiment evaluates a subsequent, separately fine-tuned policy.
Simulation predicts future video under prescribed skeleton controls and camera poses.
Policy evaluation combines action prediction on recorded trajectories with closed-loop execution on the deployment robot, separating offline prediction quality from physical task completion.
Simulation uses $480\times832$ images (height $\times$ width).
External baselines assess overall policy performance, while internal ablations examine conditioning design, simulator adaptation, and the policy-training data recipe.

\paragraph{Evaluation benchmarks.}
Simulation and offline policy evaluation use the same evaluation dataset, comprising 19 subsets collected separately from the training data.
Each episode belongs to one subset, and none of the source datasets is used for training.
The in-distribution (ID) subsets contain pick-and-place interactions representative of robot training data.
The out-of-distribution (OOD) subsets evaluate changes in visual conditions, including scene appearance, object poses, and distractors, as well as variation in objects, language instructions, spatial arrangements, task composition, and action primitives.
OOD therefore includes both familiar behaviors under changed conditions and behaviors absent from robot demonstrations; some of the latter are covered by human training data.
The subsets retain their original collection-based grouping rather than being repartitioned into uniformly task- or scene-disjoint splits.
We report simulation results separately on ID and OOD subsets, and aggregate offline policy results across all 19 subsets under the label \emph{Benchmark}.

\paragraph{Real-robot evaluation.}
We evaluate policies on the IRON-R01-1.11 humanoid, measuring task-progress scores and success rates on selected tasks with 20 trials per task per method under matched resets.
We first present simulation fidelity and rollout efficiency (\S\ref{sec:exp-sim}), followed by real-robot policy performance, human-to-robot skill transfer, training-data ablations, generalization analysis, and self-improvement (\S\ref{sec:exp-policy}).

\paragraph{Evaluation questions.}
The experiments address six questions:
(Q1) Can the simulator predict controllable futures and remain stable under autoregressive rollouts?
(Q2) How well does the complete policy perform on real-robot ID, task-level OOD, and distractor-robustness tests?
(Q3) Can manipulation skills learned from human data transfer to the robot on tasks absent from robot demonstrations?
(Q4) What do video pretraining and continued human supervision contribute under matched adaptation settings?
(Q5) How do these benefits vary with human and robot behavioral coverage?
(Q6) Can simulation improve policy development through recovery supervision
and qualitative inspection of policy-conditioned outcomes?

\subsection{Simulation}
\label{sec:exp-sim}

We evaluate video prediction under prescribed skeleton controls and camera poses on the ID and OOD subsets of the evaluation dataset (\S\ref{sec:exp-setup}).
We first compare skeleton-conditioning strategies, then assess whether self-gradient forcing (SGF) improves autoregressive prediction for subsequent policy evaluation and recovery-data synthesis.
We report PSNR and SSIM for image fidelity, Optical Flow Similarity (OFS)\footnote{OFS measures the average cosine similarity between optical flow fields computed using OpenCV Farneback from predicted and ground-truth videos.} for motion consistency, and DINO feature similarity\footnote{DINO feature similarity is computed using a DINOv3-large encoder.} for visual feature agreement.
Higher values indicate better quality for all four metrics; latency and memory measure inference cost.

\subsubsection{Short-horizon prediction: skeleton conditioning}
\label{sec:exp-fk}
\label{sec:exp-actcond}

\paragraph{Setup.}
All four variants are initialized from Wan2.2-TI2V-5B~\citep{wan2025} and trained on L4$+$F with teacher forcing.
They receive skeleton control latents aligned with the RGB video latents and are evaluated on 97-frame videos with 50 ODE sampling steps (Table~\ref{tab:act_cond}).

\paragraph{Conditioning strategies.}
\textbf{Channel concatenation} joins RGB and skeleton latents before patch embedding; \textbf{token addition} embeds them separately and adds aligned tokens at the input.
\textbf{AdaLN} uses skeleton tokens to modulate self-attention and feed-forward features in each DiT block, while \textbf{cross-attention} lets video tokens attend to skeleton tokens within the same latent frame in each block.

\begin{table}[t]
  \centering
  \caption{\textbf{Skeleton-conditioning ablation.}
  Evaluation on 97-frame videos with 50 denoising steps.
  Bold marks the best quality metric within each split.
  Latency is in seconds and memory in GB; measurements use four significant figures.}
  \label{tab:act_cond}
  \footnotesize
  \setlength{\tabcolsep}{2.5pt}
  \renewcommand{\arraystretch}{1.15}
  \resizebox{\linewidth}{!}{%
  \begin{tabular}{@{}lcccccc@{\hspace{10pt}}cccccc@{}}
    \toprule
    & \multicolumn{6}{c}{\textbf{In-distribution}}
    & \multicolumn{6}{c}{\textbf{Out-of-distribution}} \\
    \cmidrule(lr){2-7}\cmidrule(l){8-13}
    Injection
      & PSNR$\uparrow$ & SSIM$\uparrow$ & OFS$\uparrow$ & DINO$\uparrow$
      & Latency$\downarrow$ & Mem.$\downarrow$
      & PSNR$\uparrow$ & SSIM$\uparrow$ & OFS$\uparrow$ & DINO$\uparrow$
      & Latency$\downarrow$ & Mem.$\downarrow$ \\
    \midrule
    AdaLN
      & 17.12 & 0.5818 & \textbf{0.3864} & 0.9770 & 61.25 & 41.41
      & 16.88 & 0.5818 & 0.4500 & 0.9678 & 59.68 & 41.41 \\
    Cross-attention
      & 16.29 & 0.5631 & 0.3308 & 0.9764 & 59.60 & 41.27
      & 16.31 & 0.5703 & 0.3723 & 0.9688 & 59.50 & 41.27 \\
    Channel concat.
      & 17.62 & 0.5902 & 0.3760 & 0.9732 & 57.76 & 41.27
      & 17.47 & 0.5865 & 0.4565 & 0.9646 & 57.13 & 41.27 \\
    Token addition
      & \textbf{18.07} & \textbf{0.6030} & 0.3854 & \textbf{0.9809} & 57.68 & 41.27
      & \textbf{17.99} & \textbf{0.6109} & \textbf{0.4632} & \textbf{0.9765} & 56.96 & 41.27 \\
    \bottomrule
  \end{tabular}}
\end{table}

\paragraph{Results.}
Token addition achieves the highest PSNR, SSIM, and DINO scores on both splits and the highest OOD OFS, with ID OFS slightly below AdaLN.
It improves PSNR over AdaLN by $0.95$\,dB on ID and $1.11$\,dB on OOD without increasing inference cost, supporting its use as the default.

\subsubsection{Long-horizon rollout: self-gradient forcing}
\label{sec:exp-rollout}
\label{sec:exp-rollout-proto}

\paragraph{Setup.}
We further train the token-addition baseline with SGF on L4$+$F, following Stage~III-sim (\S\ref{sec:stage-iii-sim}); this model is denoted \textbf{Baseline + SGF}.
Both models generate video autoregressively from an initial observed frame under prescribed controls.
The baseline uses 50 ODE sampling steps per chunk at inference, while the SGF-trained model uses 8 to match the sampling schedule used during SGF training.
Evaluation covers 97-frame videos (100 ID and 52 OOD) and 193-frame videos (22 ID and 11 OOD), corresponding to six and twelve prediction chunks, respectively (Table~\ref{tab:rollout_ablate}).

\begin{table}[t]
  \centering
  \caption{\textbf{Effect of self-gradient forcing on video prediction.}
  Baseline + SGF is the Token addition baseline further trained with SGF.
  The baseline uses 50 ODE sampling steps per chunk; the SGF-trained baseline uses 8.
  Bold marks the best quality metric within each horizon and split.
  Latency is in seconds and memory in GB; measurements use four significant figures.}
  \label{tab:rollout_ablate}
  \label{tab:rollout_steps}
  \label{tab:rollout_denoise}
  \footnotesize
  \setlength{\tabcolsep}{2.5pt}
  \renewcommand{\arraystretch}{1.15}
  \resizebox{\linewidth}{!}{%
  \begin{tabular}{@{}lcccccc@{\hspace{10pt}}cccccc@{}}
    \toprule
    & \multicolumn{6}{c}{\textbf{In-distribution}}
      & \multicolumn{6}{c}{\textbf{Out-of-distribution}} \\
    \cmidrule(lr){2-7}\cmidrule(l){8-13}
    Method
      & PSNR$\uparrow$ & SSIM$\uparrow$ & OFS$\uparrow$ & DINO$\uparrow$
      & Latency$\downarrow$ & Mem.$\downarrow$
      & PSNR$\uparrow$ & SSIM$\uparrow$ & OFS$\uparrow$ & DINO$\uparrow$
      & Latency$\downarrow$ & Mem.$\downarrow$ \\
    \midrule
    \multicolumn{13}{l}{\textit{Short-horizon (97 frames)}} \\
    Baseline
      & 18.07 & 0.6030 & 0.3854 & 0.9809 & 57.68 & 41.27
      & 17.99 & 0.6109 & 0.4632 & 0.9765 & 56.96 & 41.27 \\
    Baseline + SGF
      & \textbf{19.17} & \textbf{0.6394} & \textbf{0.5696} & \textbf{0.9848} & 20.95 & 41.27
      & \textbf{18.72} & \textbf{0.6347} & \textbf{0.5467} & \textbf{0.9799} & 20.00 & 41.27 \\
    \midrule
    \multicolumn{13}{l}{\textit{Long-horizon (193 frames)}} \\
    Baseline
      & 16.94 & 0.5651 & 0.4035 & 0.9782 & 119.7 & 44.96
      & 17.74 & 0.6009 & 0.4878 & 0.9755 & 120.2 & 44.96 \\
    Baseline + SGF
      & \textbf{17.83} & \textbf{0.5976} & \textbf{0.4797} & \textbf{0.9825} & 39.86 & 44.94
      & \textbf{18.75} & \textbf{0.6350} & \textbf{0.5563} & \textbf{0.9797} & 36.44 & 44.96 \\
    \bottomrule
  \end{tabular}}
\end{table}

\paragraph{Results.}
The SGF-trained baseline improves all four quality metrics on both splits at both horizons.
At 193 frames, PSNR increases by $0.89$\,dB on ID and $1.01$\,dB on OOD, with $3.00\times$ and $3.30\times$ speedups and essentially unchanged memory use.
The consistent gains in image fidelity and motion consistency across ID and OOD videos demonstrate the effectiveness of SGF for autoregressive prediction.

\begin{wraptable}{R}{0.50\textwidth}
  \centering
  \captionsetup{font=small,skip=5pt}
  \caption{\textbf{Final simulator (Phase~II-c + SGF).}}
  \label{tab:dagger_simulator}
  \small
  \setlength{\tabcolsep}{4pt}
  \renewcommand{\arraystretch}{1.1}
  \begin{tabular}{@{}lcccc@{}}
    \toprule
    Frames & PSNR$\uparrow$ & SSIM$\uparrow$ & OFS$\uparrow$ & DINO$\uparrow$ \\
    \midrule
    \multicolumn{5}{l}{\textit{In-distribution}} \\
    97  & 19.73 & 0.6612 & 0.6007 & 0.9853 \\
    193 & 18.11 & 0.6086 & 0.5170 & 0.9834 \\
    \midrule
    \multicolumn{5}{l}{\textit{Out-of-distribution}} \\
    97  & 18.44 & 0.6274 & 0.5642 & 0.9758 \\
    193 & 18.05 & 0.6142 & 0.5552 & 0.9705 \\
    \bottomrule
  \end{tabular}
\end{wraptable}

\paragraph{Simulator used for recovery-data synthesis.}
For recovery-data synthesis, we apply Stage~III-sim SGF fine-tuning to the full \modelname checkpoint after Phase~II-c (\S\ref{sec:stage-ii}), rather than to the L4$+$F ablation baseline.
Table~\ref{tab:dagger_simulator} reports this final simulator's prediction quality, averaged over the same evaluation videos at both horizons.
The simulator achieves excellent prediction quality on both ID and OOD videos, supporting its use for recovery-data synthesis.
We then use this simulator to generate the DAgger data for the real-robot self-improvement experiment in \S\ref{sec:policy-self-improvement}.

\Needspace{4\baselineskip}
\subsection{Policy}
\label{sec:exp-policy}

\paragraph{Evaluation suite.}
Our primary policy benchmark consists of three real-robot manipulation tasks executed on the deployment humanoid:
banana pick-and-place across five fixed object positions, pouring water
from a bottle, and bowl stacking.  Each method is evaluated for
20 trials per task under matched resets.
Tasks~1 and~2 are in-distribution (ID), whereas Task~3 evaluates task-level out-of-distribution (OOD) generalization relative to L4 robot demonstrations.
Bowl stacking is absent from L4 but present in L2 human data and bridge
data.  Task~3 therefore tests transfer of human-observed skills to robot
execution, not a task unseen across all training sources.
Task~4, introduced separately
below, adds distractor objects and is treated as a robustness experiment
rather than part of the primary three-task benchmark.
Because a binary outcome does not distinguish early failure from partial
completion,
we report two metrics per task: \textbf{success rate} (SR), and a
dense \textbf{task-progress score} that awards partial credit at
fixed milestones (e.g., reach, grasp, transport, place; per-task
rubrics in~\S\ref{sec:exp-ext-ood}).
We compare against two external baselines under the same rollout
protocol, embodiment, and trial budget: \textbf{GR00T}~\citep{grootn1},
a VLA policy, and \textbf{DreamZero}~\citep{dreamzero2026}, a
world-model-based policy.
All methods are trained on the same robot teleoperation corpus and the same action-supervised human data; the video-only adaptation in Stage~I is used by \modelname{} alone.

\paragraph{Comparison and attribution protocol.}
The external comparison varies the complete model and training recipe and
therefore measures end-to-end performance.  Attribution instead uses the
internal controls below.  We first compare robot-only and full-data training
in closed loop, then examine Stage~I exposure under identical Stage~II
fine-tuning and human co-training under matched initialization and budget.
Coverage-stratified analysis identifies where the human-data gains occur.
Finally, we evaluate policy refinement using recovery data from the fixed
simulator.  This order separates overall performance, training-recipe
contributions, the distribution of transfer gains, and self-improvement.

\subsubsection{Real-robot performance and robustness}
\label{sec:exp-ext-ood}

\paragraph{Tasks and progress rubrics.}
Milestones and success criteria were fixed before evaluation.
\textbf{Task~1 --- banana pick-and-place (ID; position robustness):}
``pick up the banana and place it in the basket.''  The banana is
placed at five predefined workspace locations, with four trials per
location.  Every method uses the same positions and resets for a fair
comparison, so the task measures position robustness under controlled
initial conditions rather than robustness to randomly sampled positions.
Credit: $0.25$ hand touches the banana, $0.5$ banana lifted, $0.75$
banana touches the basket, $1.0$ banana placed inside (success).
\textbf{Task~2 --- pour water (ID):} ``pour the water out of the bottle.''
Credit in three equal milestones: hand touches the bottle, bottle
lifted, water poured (bottle rotated past $45^\circ$; success).  This
task stresses sustained wrist rotation with an in-hand object.
\textbf{Task~3 --- stack bowls (task OOD):} ``stack the bowls together.''
Credit: $0.25$ hand touches a bowl, $0.5$ bowl lifted, $0.75$ bowl
transported to contact the other bowl, $1.0$ bowls stacked (success).
This task stresses rim alignment and precise release.
The progress score is the per-trial credit averaged over the 20
trials; SR counts only full completions.

\begin{figure}[!htb]
  \centering
  \resizebox{\linewidth}{!}{\begin{tikzpicture}[
  font=\fontfamily{phv}\selectfont\scriptsize,
  note/.style={font=\fontfamily{phv}\selectfont\tiny, text=xpengblack, align=center},
  val/.style={font=\fontfamily{phv}\selectfont\tiny, text=xpengblack, inner sep=1pt},
]
  \fill[xpengavgband, rounded corners=3pt]
    (4.78,-0.50) rectangle (6.08,3.38);
  \foreach \y/\lab in {0/0,0.8/.25,1.6/.50,2.4/.75,3.2/1} {
    \draw[xpenggrid, densely dashed, line width=0.5pt]
      (0,\y) -- (6.1,\y);
    \node[note, anchor=east] at (-0.12,\y) {\lab};
  }
  \draw[xpengblack, line width=0.65pt] (0,0) -- (6.15,0);
  \draw[xpengblack, line width=0.65pt] (0,0) -- (0,3.30);
  \foreach \x/\a/\b/\c/\la/\lb/\lc in {
    0.60/1.28/2.64/2.96/.40/.83/.93,
    2.00/1.07/1.39/2.51/.33/.43/.78,
    3.40/1.12/2.48/2.64/.35/.78/.83,
    5.00/1.16/2.17/2.70/.36/.68/.84
  } {
    \fill[xpengbargold]  (\x,0)      rectangle ++(0.30,\a);
    \fill[xpengbargreen] (\x+0.34,0) rectangle ++(0.30,\b);
    \fill[xpengbarblue]  (\x+0.68,0) rectangle ++(0.30,\c);
    \node[val, anchor=south] at (\x+0.15,\a) {\la};
    \node[val, anchor=south] at (\x+0.49,\b) {\lb};
    \node[val, anchor=south] at (\x+0.83,\c) {\lc};
  }
  \node[note] at (1.09,-0.30) {T1 banana};
  \node[note] at (2.49,-0.30) {T2 pour};
  \node[note] at (3.89,-0.30) {T3 stack};
  \node[note] at (5.49,-0.30) {Avg};
  \node[font=\fontfamily{phv}\selectfont\scriptsize\bfseries, anchor=south] at (3.075,3.42)
    {(a) Task-progress score};

  \begin{scope}[xshift=7.4cm]
    \fill[xpengavgband, rounded corners=3pt]
      (4.78,-0.50) rectangle (6.08,3.38);
    \foreach \y/\lab in {0/0,0.8/25,1.6/50,2.4/75,3.2/100} {
      \draw[xpenggrid, densely dashed, line width=0.5pt]
        (0,\y) -- (6.1,\y);
      \node[note, anchor=east] at (-0.12,\y) {\lab};
    }
    \draw[xpengblack, line width=0.65pt] (0,0) -- (6.15,0);
    \draw[xpengblack, line width=0.65pt] (0,0) -- (0,3.30);
    \foreach \x/\a/\b/\c/\la/\lb/\lc in {
      0.60/0.64/2.08/2.56/20/65/80,
      2.00/0.001/0.16/1.60/0/5/50,
      3.40/0.001/1.60/2.40/0/50/75,
      5.00/0.21/1.28/2.19/7/40/68
    } {
      \fill[xpengbargold]  (\x,0)      rectangle ++(0.30,\a);
      \fill[xpengbargreen] (\x+0.34,0) rectangle ++(0.30,\b);
      \fill[xpengbarblue]  (\x+0.68,0) rectangle ++(0.30,\c);
      \node[val, anchor=south] at (\x+0.15,\a) {\la};
      \node[val, anchor=south] at (\x+0.49,\b) {\lb};
      \node[val, anchor=south] at (\x+0.83,\c) {\lc};
    }
    \node[note] at (1.09,-0.30) {T1 banana};
    \node[note] at (2.49,-0.30) {T2 pour};
    \node[note] at (3.89,-0.30) {T3 stack};
    \node[note] at (5.49,-0.30) {Avg};
    \node[font=\fontfamily{phv}\selectfont\scriptsize\bfseries, anchor=south] at (3.075,3.42)
      {(b) Success rate (\%)};
  \end{scope}

  \draw[xpenggrid, fill=white, rounded corners=1pt, line width=0.45pt]
    (3.55,4.02) rectangle (10.25,4.44);
  \fill[xpengbargold] (4.05,4.15) rectangle ++(0.28,0.16);
  \node[note, anchor=west] at (4.45,4.23) {GR00T};
  \fill[xpengbargreen] (5.92,4.15) rectangle ++(0.28,0.16);
  \node[note, anchor=west] at (6.32,4.23) {DreamZero};
  \fill[xpengbarblue] (8.20,4.15) rectangle ++(0.28,0.16);
  \node[note, anchor=west] at (8.60,4.23) {Ours};
\end{tikzpicture}}
  \caption{Real-robot policy evaluation (measured): (a) task-progress
  score and (b) success rate on the three-task suite (ID:
  Tasks~1--2; task-level OOD: Task~3), 20 trials per task per method.
  \modelname{} leads on every task under
  both metrics.  Progress credits intermediate milestones, while success
  requires completing the entire manipulation.}
  \label{fig:policy_eval_suite}
\end{figure}
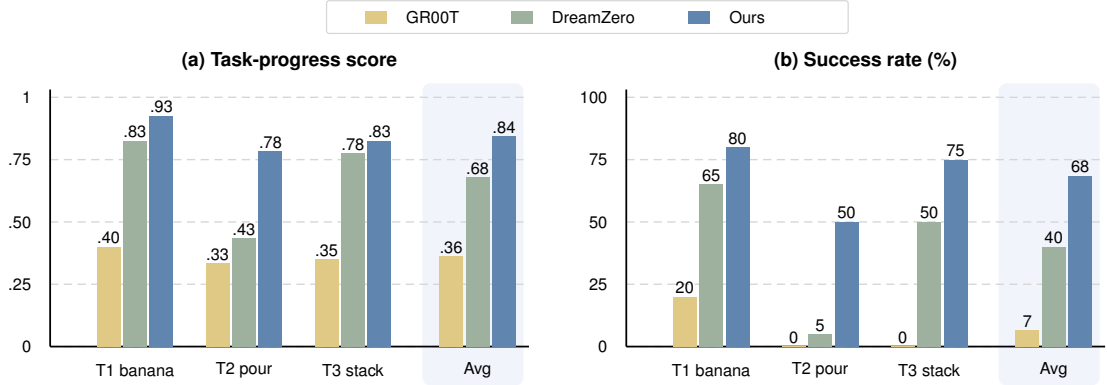

\begin{wrapfigure}{R}{0.40\textwidth}
  \centering
  \captionsetup{font=small,skip=5pt}
  \resizebox{0.90\linewidth}{!}{\begin{tikzpicture}[
  y=0.75cm,
  font=\fontfamily{phv}\selectfont\scriptsize,
  note/.style={font=\fontfamily{phv}\selectfont\tiny, text=xpengblack, align=center},
  val/.style={font=\fontfamily{phv}\selectfont\tiny, text=xpengblack, inner sep=1pt},
]
  \foreach \y/\lab in {0/0,0.8/.25,1.6/.50,2.4/.75,3.2/1} {
    \draw[xpenggrid, densely dashed, line width=0.5pt]
      (0,\y) -- (3.9,\y);
    \node[note, anchor=east] at (-0.12,\y) {\lab};
  }
  \draw[xpengblack, line width=0.65pt] (0,0) -- (3.95,0);
  \draw[xpengblack, line width=0.65pt] (0,0) -- (0,3.30);
  \foreach \x/\a/\b/\la/\lb in {
    0.70/2.24/2.98/.70/.93,
    2.40/1.54/2.37/.48/.74
  } {
    \fill[xpengbargold]  (\x,0)      rectangle ++(0.42,\a);
    \fill[xpengbarblue]  (\x+0.50,0) rectangle ++(0.42,\b);
    \node[val, anchor=south] at (\x+0.21,\a) {\la};
    \node[val, anchor=south] at (\x+0.71,\b) {\lb};
  }
  \node[note] at (1.16,-0.30) {T1 positions};
  \node[note] at (2.86,-0.30) {T4 distractor};
  \node[rotate=90, font=\fontfamily{phv}\selectfont\scriptsize\bfseries] at (-0.80,1.60) {(a) Task-progress score};

  \begin{scope}[yshift=-3.1875cm]
    \foreach \y/\lab in {0/0,0.8/25,1.6/50,2.4/75,3.2/100} {
      \draw[xpenggrid, densely dashed, line width=0.5pt]
        (0,\y) -- (3.9,\y);
      \node[note, anchor=east] at (-0.12,\y) {\lab};
    }
    \draw[xpengblack, line width=0.65pt] (0,0) -- (3.95,0);
    \draw[xpengblack, line width=0.65pt] (0,0) -- (0,3.30);
    \foreach \x/\a/\b/\la/\lb in {
      0.70/1.60/2.56/50/80,
      2.40/0.96/1.60/30/50
    } {
      \fill[xpengbargold]  (\x,0)      rectangle ++(0.42,\a);
      \fill[xpengbarblue]  (\x+0.50,0) rectangle ++(0.42,\b);
      \node[val, anchor=south] at (\x+0.21,\a) {\la};
      \node[val, anchor=south] at (\x+0.71,\b) {\lb};
    }
    \node[note] at (1.16,-0.30) {T1 positions};
    \node[note] at (2.86,-0.30) {T4 distractor};
    \node[rotate=90, font=\fontfamily{phv}\selectfont\scriptsize\bfseries] at (-0.80,1.60) {(b) Success rate (\%)};
  \end{scope}

  \draw[xpenggrid, fill=white, rounded corners=1pt, line width=0.45pt]
    (-0.05,3.68) rectangle (3.95,4.10);
  \fill[xpengbargold] (0.15,3.81) rectangle ++(0.28,0.16);
  \node[note, anchor=west] at (0.55,3.89) {Robot-only};
  \fill[xpengbarblue] (2.20,3.81) rectangle ++(0.28,0.16);
  \node[note, anchor=west] at (2.60,3.89) {Full data};
\end{tikzpicture}}
  \caption{\textbf{Closed-loop human-data ablation.}
  (a) Task-progress score and (b) success rate for the robot-only and
  full-data models on position and distractor robustness, with 20 trials per task.}
  \label{fig:policy_success_reliability}
\end{wrapfigure}
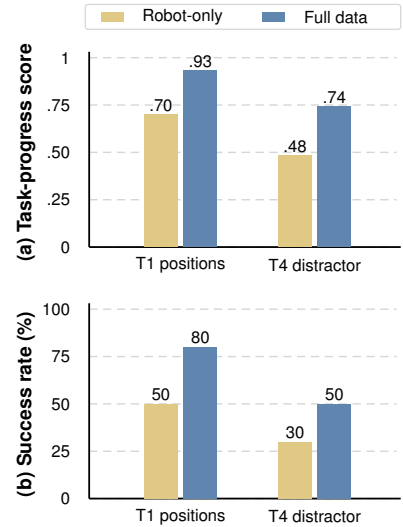

\paragraph{Results.}
\modelname{} leads on every task under both metrics
(Figure~\ref{fig:policy_eval_suite}):
$68.3\%$ average SR vs.\ $40.0\%$ for DreamZero and $6.7\%$ for GR00T,
with average progress $0.84$ vs.\ $0.68$ and $0.36$.
The comparison measures the complete model and training recipe: supervised
corpora are matched, but \modelname{} additionally receives Stage~I video
adaptation.  It should not be read as an architecture-only comparison.

The task breakdown is more informative than the average alone.  Pouring
has the lowest success rate for every method and the largest success-rate
gap between \modelname{} and DreamZero ($10/20$ versus $1/20$).  Unlike
transporting an object to a receptacle, it requires sustained rotation
while maintaining an in-hand grasp.  Progress also rises from $0.43$ to
$0.78$, indicating an improvement beyond binary completion alone.
Nevertheless, $0.78$ progress alongside $50\%$ success shows that reliable
completion remains a challenge even for the full model.  Aggregate scores
distinguish partial progress from completion, but do not by themselves
identify which motion caused a failure.

On banana pick-and-place, \modelname{} succeeds in $16/20$ trials with
successes at all five tested positions; on task-level OOD bowl stacking,
it succeeds in $15/20$.  These results motivate a more targeted question:
does the broader training corpus help when the robot encounters spatial
variation or distracting objects, beyond the advantage of the complete
recipe over external baselines?

\paragraph{Human experience improves policy robustness.}
Broader human and video experience is intended to support not only skill
transfer, but also robust execution when spatial arrangements and surrounding
objects vary.  We test this in closed loop by comparing the
\textbf{robot-only variant}, trained without human data or unsupervised
video, with the \textbf{full-data model}.  Task~1 evaluates banana
pick-and-place across five fixed positions.  A separate distractor test
(Task~4) asks the robot to ``pick up the block and put it in the box''
with a cucumber, a flower, and a toy duck---objects absent from the robot
training corpus---placed in the scene.  Each model receives 20 trials per
task, scored using the same four-milestone pick-and-place rubric.

The full-data model improves both spatial and distractor robustness
(Figure~\ref{fig:policy_success_reliability}).  Across the five positions,
success rises from $50\%$ to $80\%$ and progress from $0.70$ to $0.93$.
With unseen distractors, success rises from $30\%$ to $50\%$ and progress from $0.48$ to $0.74$.
The gains therefore extend beyond intermediate milestones to more completed trials, including in an in-distribution task family.
These results support the value of broader experience for executing
familiar manipulation skills under spatial and visual variation, not only
for transferring behaviors absent from robot demonstrations.

This comparison evaluates human action supervision and unsupervised video
learning as a combined data recipe.  The following controlled studies
(\S\ref{sec:exp-data} and~\S\ref{sec:exp-sim-cotrain}) examine video
pretraining and continued human supervision separately through offline
action prediction.

\subsubsection{Human-to-robot skill transfer}

\begin{figure}[!tb]
    \centering
    \includegraphics[width=\linewidth]{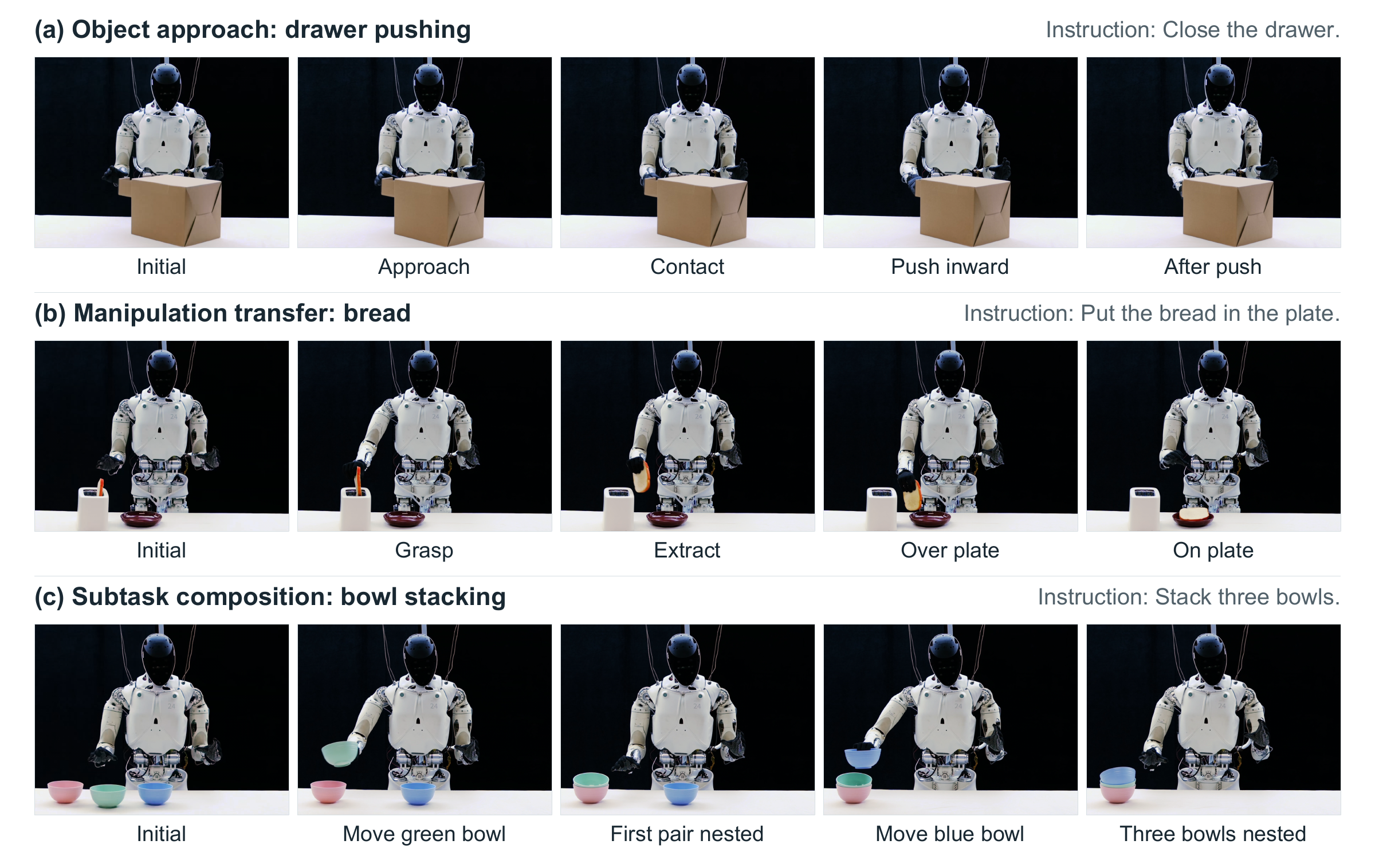}
    \caption{Real-robot transfer from human experience (qualitative).
    Each row shows five keyframes of \modelname{} following the displayed instruction.
    (a) Drawer pushing illustrates approach and contact with an object absent from robot training data.
    (b) Bread transfer adapts manipulation of an object present in both human and robot data to a toaster-to-plate task absent from robot demonstrations.
    (c) Three-bowl stacking composes two successive placements under a single task instruction.
    These examples illustrate a progression from object approach to manipulation transfer and sequential subtask composition.}
    \label{fig:progressive_transfer}
\end{figure}

\paragraph{Qualitative transfer.}
We examine three real-robot tasks that probe progressively richer forms of transfer from human experience: approaching an object absent from robot training, manipulating a familiar object with different source and target receptacles, and composing successive manipulations under a single task instruction. Figure~\ref{fig:progressive_transfer} shows representative executions at these three levels.

\paragraph{Assessment criteria.}
Evaluation follows task-specific milestones and visible outcomes. For drawer pushing, we assess target-directed approach, contact, and inward displacement. Bread transfer requires extracting the slice from the toaster and releasing it onto the plate. Bowl nesting requires completing both placements in sequence, with all three bowls remaining nested after the final release.

\paragraph{Results.}
Drawers are entirely absent from the robot training corpus. Nevertheless, \modelname{} approaches the drawer, establishes contact, and pushes it inward, illustrating transfer of approach-and-contact behavior from human experience to an unseen object. Bread appears in both human and robot data, but the toaster-to-plate task is absent from robot demonstrations. Successful extraction and placement extend transfer from object approach to familiar-object manipulation under a changed source--target configuration. Finally, under a single bowl-nesting instruction, the robot places the green bowl inside the pink bowl and then places the blue bowl into the resulting pair. The second placement builds on the state produced by the first, demonstrating sequential subtask composition within one task. Together, these examples progress from approaching an unseen object, through transferring object manipulation across receptacle configurations, to composing multiple manipulation steps toward a single goal.

\subsubsection{Contribution of video pretraining}
\label{sec:exp-data}

\paragraph{Setup.}
The closed-loop comparison above establishes the benefit of the full data recipe.
We next examine its video-only pretraining component: does Stage~I pretraining improve subsequent policy learning, and how does the benefit change with cumulative video exposure?
We take pretrained checkpoints along a single Stage~I trajectory corresponding to fractions $r\in\{0,\tfrac18,\tfrac14,\tfrac12,1\}$ of the Stage~I video corpus consumed ($r{=}0$ retains the pretrained video-DiT initialization but receives no Stage~I adaptation), then run an \emph{identical} supervised action fine-tuning recipe from each, with the action head freshly initialized from the same seed in every arm so the backbone is the only variable.
We report teacher-forced action evaluation loss (MSE in the normalized
action space, using recorded context for each chunk) on the shared
evaluation dataset described in~\S\ref{sec:exp-setup}.  Every arm is
evaluated on the same episodes, with results aggregated across the suite's
mixed distributional coverage.

\Needspace{15\baselineskip}
\begin{wrapfigure}[17]{r}{0.48\textwidth}
  \centering
  \resizebox{\linewidth}{!}{\begin{tikzpicture}[font=\fontfamily{phv}\selectfont\scriptsize]
  \node[anchor=west,font=\fontfamily{phv}\selectfont\scriptsize\bfseries] at (-0.55,3.05)
    {Benchmark action-loss change ($\downarrow$)};
  \foreach \y/\label in {2.60/0,1.70/-5,0.80/-10} {
    \draw[black!15] (0,\y) -- (4.95,\y);
    \node[anchor=east,text=xpengmuted] at (-0.12,\y) {\label\%};
  }
  \draw[black!50] (0,0.10) -- (4.95,0.10);
  \foreach \x/\label in {0/0,1.2/{1/8},2.4/{1/4},3.6/{1/2},4.8/1} {
    \draw[black!50] (\x,0.10) -- (\x,0.02);
    \node[anchor=north] at (\x,-0.06) {\label};
  }
  \node[font=\fontfamily{phv}\selectfont\scriptsize\bfseries] at (2.4,-0.62) {Stage~I exposure fraction \textit{r}};
  \draw[xpengblue,line width=1.1pt] plot coordinates {
    (0,2.60) (1.2,1.142) (2.4,0.98) (3.6,0.98) (4.8,0.35)};
  \foreach \x/\y in {0/2.60,1.2/1.142,2.4/0.98,3.6/0.98,4.8/0.35}
    \fill[xpengblue] (\x,\y) circle (1.8pt);
  \foreach \x/\y/\label in {1.2/1.142/-8.1,2.4/0.98/-9.0,3.6/0.98/-9.0,4.8/0.35/-12.5}
    \node[above=4pt,text=xpengblue] at (\x,\y) {\label\%};
\end{tikzpicture}}
  \captionsetup{font=small}
  \caption{Stage~I exposure and benchmark action loss relative to
  $r{=}0$, after matched fine-tuning (tail-window average).}
  \label{fig:human_ratio}
\end{wrapfigure}
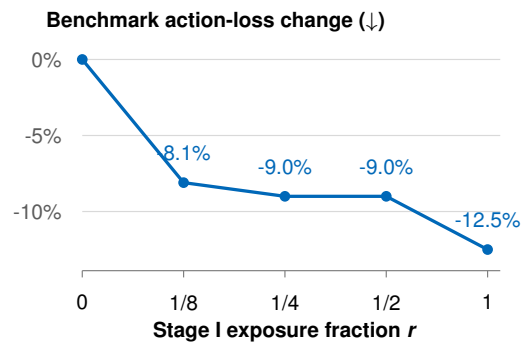

\paragraph{Results and implications.}
Stage~I improves policy learning under the fixed adaptation budget: every nonzero exposure reduces action loss on the evaluation dataset, as shown in Figure~\ref{fig:human_ratio}.
The first eighth of exposure already captures approximately two-thirds of the full-exposure benchmark reduction ($8.1\%$ versus $12.5\%$).
The reduction reaches $9.0\%$ at both quarter and half exposure, and $12.5\%$ with full Stage~I exposure.
Thus, modest exposure provides a substantial early benefit, while full
exposure gives the best aggregate result in this sweep.

The benefit appears after every backbone receives the same action-training
recipe.  Video prediction therefore provides useful preparation for control
even without action labels, supporting the separation of broad predictive
learning from action grounding in~\S\ref{sec:training}.  Robot supervision
need not supply all of the experience that makes action prediction effective.
This comparison measures the value of additional Stage~I training: corpus exposure and optimization compute increase together.

\subsubsection{Human supervision during robot adaptation}
\label{sec:exp-sim-cotrain}

\paragraph{Should human supervision end when robot adaptation begins?}
Video-only adaptation tests whether visual experience benefits the policy
backbone (\S\ref{sec:exp-data}).  Here we ask whether human action
supervision should serve only as an initialization or remain available
while the model learns robot execution.  We compare three recipes on the
held-out action benchmark: a robot-only baseline; human--robot mid-training
followed by robot-only fine-tuning (mid-train$\rightarrow$FT);
human--robot \emph{co-training} kept through the final stage, including
the bridge-mined subsets.  The two human-data recipes share one mid-trained
initialization, learning rate, and budget, so the
mid-train$\rightarrow$FT vs.\ co-train comparison isolates a single
variable: whether human data stays in the mixture during the final
stage.  Table~\ref{tab:human_supervision_ablation} reports changes relative to the
robot-only baseline, averaged over nine late-training checkpoints.

\begin{table}[!htb]
  \centering
  \caption{Supervised human data: held-out action-loss change vs.\ a robot-only baseline (mean over nine checkpoints; across-checkpoint s.d.\ in percentage points in parentheses).
  The checkpoint spread describes variation within a training run, not uncertainty across independent seeds.
  Co-training gives the larger mean reduction under a matched initialization.}
  \label{tab:human_supervision_ablation}
  \small
  \begin{tabular}{@{}lc@{}}
    \toprule
    Recipe & Benchmark $\Delta$ action loss$\downarrow$ \\
    \midrule
    Robot-only & $0$ (ref.) \\
    Human mid-training then robot-only FT
      & $-8.5\%$ (s.d.\ $2.7$) \\
    Human--robot co-training & $\mathbf{-14.0\%}$ (s.d.\ $2.8$) \\
    \bottomrule
  \end{tabular}
\end{table}

\paragraph{Human experience as a regularizer during robot adaptation.}
With the same mid-trained initialization and final-stage budget,
co-training reduces mean benchmark action loss by $14.0\%$ relative to
robot-only, compared with $8.5\%$ for human mid-training followed by
robot-only fine-tuning (Table~\ref{tab:human_supervision_ablation}).
Because the two human-data arms begin from the same checkpoint, their gap
shows that the benefit of human experience is not exhausted by mid-training.
Its continued presence in the adaptation mixture matters for the final
action predictor.

A plausible explanation is that continued human supervision constrains
how the shared backbone specializes.  Robot-only updates reward performance
on a narrower set of scenes and behaviors, but do not require the model to
retain capabilities supported only by earlier human experience.  This
creates a risk of overfitting to robot-specific correlations and forgetting
previously learned behaviors.  Human co-training keeps those behaviors
under active supervision: human action targets continue to train motion
prediction, while paired video targets encourage the backbone to retain
visual-dynamics knowledge beyond the robot corpus.  In this interpretation,
human data acts as rehearsal during adaptation, allowing robot execution
to improve without leaving broader visuomotor capabilities unconstrained.

This motivates the Phase~II-c design: retain human supervision to preserve
transferable features and continue learning beyond the robot corpus while
specializing to deployment.
We next examine whether the benefit is confined to behaviors missing from
robot training or also improves those the robot has already demonstrated.

\subsubsection{Cross-embodiment generalization}
\label{sec:exp-ood-category}

\begin{figure}[!htbp]
  \centering
  \resizebox{0.96\linewidth}{!}{\begin{tikzpicture}[
  font=\fontfamily{phv}\selectfont\scriptsize,text=xpengblack,
  numeric/.style={font=\fontfamily{phv}\selectfont\scriptsize}]
  \fill[xpengbargold] (-6.0,4.90) rectangle (-5.65,5.08);
  \node[anchor=west] at (-5.50,4.99)
    {Human mid-training $\rightarrow$ robot-only FT};
  \fill[xpengbarblue] (1.30,4.90) rectangle (1.65,5.08);
  \node[anchor=west] at (1.80,4.99) {Human--robot co-training};

  \node[align=center,font=\fontfamily{phv}\selectfont\scriptsize\bfseries] at (-6.15,4.20)
    {Robot\\coverage};
  \node[align=center,font=\fontfamily{phv}\selectfont\scriptsize\bfseries] at (-4.20,4.20)
    {Human\\coverage};
  \foreach \x/\y/\label/\level in {
      -6.15/3.45/Dense/2,-4.20/3.45/Dense/2,
      -6.15/2.35/Absent/0,-4.20/2.35/Dense/2,
      -6.15/1.25/Absent/0,-4.20/1.25/Sparse/1,
      -6.15/0.15/Absent/0,-4.20/0.15/Absent/0} {
    \ifcase\level
      \def\badgefill{black!4}\def\badgeink{xpengmuted}
    \or
      \def\badgefill{xpengbargreen!14}\def\badgeink{xpengbargreen!55!black}
    \or
      \def\badgefill{xpengbargreen!35}\def\badgeink{xpengbargreen!55!black}
    \fi
    \fill[\badgefill,rounded corners=3pt]
      ({\x-0.87},{\y-0.23}) rectangle ({\x+0.87},{\y+0.23});
    \begin{scope}[shift={({\x-0.63},\y)}]
      \ifnum\level=2
        \fill[\badgeink] (0,0) circle (0.085);
      \else
        \fill[white] (0,0) circle (0.085);
        \ifnum\level=1
          \fill[\badgeink] (0,0) -- (0,0.085) arc (90:270:0.085) -- cycle;
        \fi
        \draw[\badgeink,line width=0.55pt] (0,0) circle (0.085);
      \fi
    \end{scope}
    \node[anchor=west,text=\badgeink]
      at ({\x-0.47},\y) {\label};
  }

  \foreach \v in {-10,-5,5,10,15,20,25} {
    \draw[xpenggrid,densely dashed,line width=0.5pt] ({0.25*\v},-0.35) -- ({0.25*\v},3.88);
    \node[numeric,anchor=north] at ({0.25*\v},-0.46) {\v};
  }
  \draw[xpengblack,line width=0.65pt] (0,-0.35) -- (0,3.88);
  \node[numeric,anchor=north] at (0,-0.46) {0};

  \foreach \y/\mid/\co in {
      3.45/5.1/10.0,2.35/14.9/22.7,1.25/7.3/10.8,0.15/-9.3/0.3} {
    \fill[xpengbargold] (0,{\y+0.06}) rectangle ({0.25*\mid},{\y+0.34});
    \fill[xpengbarblue] (0,{\y-0.34}) rectangle ({0.25*\co},{\y-0.06});
    \ifdim \mid pt<0pt
      \node[numeric,anchor=east] at ({0.25*\mid-0.08},{\y+0.20})
        {\mid};
    \else
      \node[numeric,anchor=west] at ({0.25*\mid+0.08},{\y+0.20})
        {\mid};
    \fi
    \node[numeric,anchor=west] at ({0.25*\co+0.08},{\y-0.20})
      {\co};
  }
  \node[font=\fontfamily{phv}\selectfont\scriptsize\bfseries] at (1.88,-1.05)
    {Action-loss reduction vs.\ robot-only (\%; higher is better)};
\end{tikzpicture}}
  \caption{\textbf{Human-to-robot transfer across training-data coverage.}
  Relative reduction in held-out action loss against robot-only, averaged
  over nine late-training checkpoints.  Positive values indicate lower
  loss; negative values indicate higher loss.  Coverage refers to the
  corresponding task family in each training source. }
  \label{fig:coverage_strata}
\end{figure}
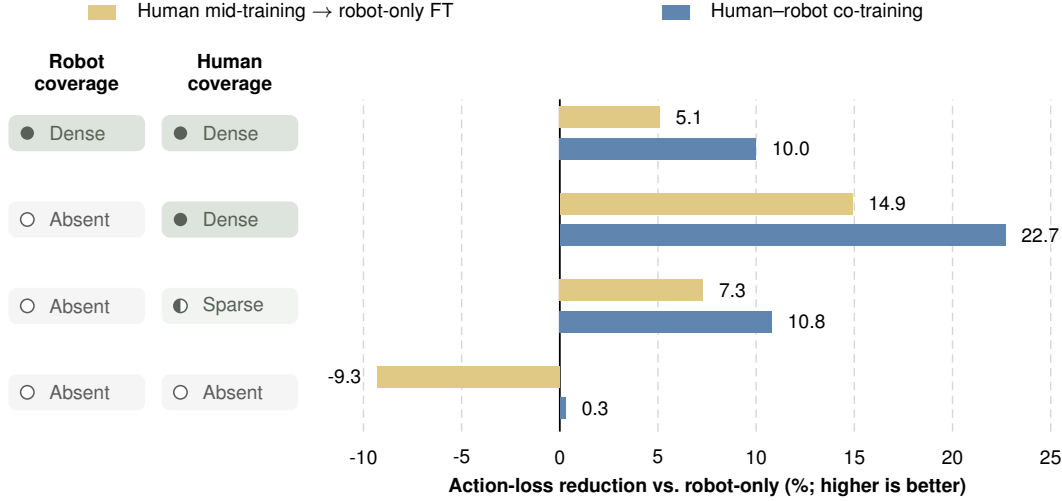

\paragraph{Coverage-stratified analysis.}
The preceding comparison establishes the aggregate benefit of human
supervision.  Here we examine how that benefit is distributed across
familiar robot behaviors and behaviors absent from robot demonstrations.
We group the 19
curated benchmark subsets into eight task-semantic families, then assign
coverage strata along two axes: whether the corresponding action family is \emph{covered by robot
teleoperation}, and whether it is \emph{densely covered by the human
corpus}.  These semantic families and coverage strata are secondary
analysis groupings, not the benchmark's original dataset partitions.
Figure~\ref{fig:coverage_strata} reports the reduction in action
evaluation loss for the two human-data recipes
of~\S\ref{sec:exp-sim-cotrain} relative to a robot-only baseline,
averaged across nine late-training checkpoints.
Stratification reveals two complementary contributions: strengthening
behaviors covered by teleoperation and extending supervision to behaviors
that robot training lacks.

\paragraph{Human supervision strengthens and extends robot capabilities.}
Human experience benefits both sides of the coverage spectrum.
Co-training reduces action loss by $10.0\%$ on robot-dense,
human-dense behaviors and by $22.7\%$ on robot-absent, human-dense
behaviors (Figure~\ref{fig:coverage_strata}).  Shared task coverage is
therefore useful, not redundant, while broader human coverage enables
transfer beyond the behaviors demonstrated by the robot.  The two
human-data recipes show the same ordering across strata, with smaller
gains under sparse human coverage and no comparable benefit when both
sources lack the behavior.  The central finding is not that human data
should replace robot demonstrations, but that robot demonstrations need
not define the full range of experience available for learning control.

\paragraph{Implications for the data curriculum.}
Together with the matched adaptation comparison, these results support retaining human supervision in Phase~II-c for both robot-covered and robot-absent behaviors.
They are consistent with the complementary data roles in \S\ref{sec:data}, but do not isolate the contribution of bridge data or establish a particular feature-alignment mechanism.
The next experiment examines a different source of additional policy supervision: recovery examples synthesized by the simulator rather than drawn from recorded human experience.

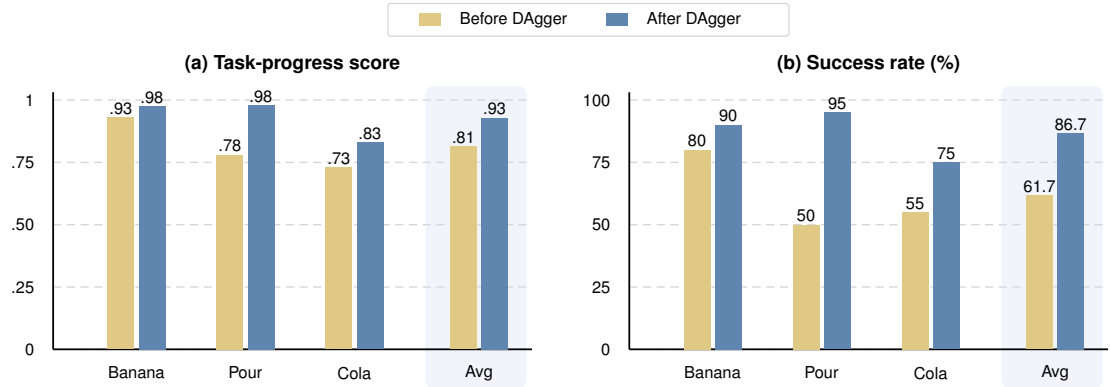
\begin{figure}[!htbp]
  \centering
  \resizebox{\linewidth}{!}{\begin{tikzpicture}[
  font=\fontfamily{phv}\selectfont\scriptsize,
  note/.style={font=\fontfamily{phv}\selectfont\tiny, text=xpengblack, align=center},
  val/.style={font=\fontfamily{phv}\selectfont\tiny, text=xpengblack, inner sep=1pt},
]
  \fill[xpengavgband, rounded corners=3pt]
    (4.78,-0.50) rectangle (6.08,3.38);
  \foreach \height/\ticklabel in {0/0,0.8/.25,1.6/.50,2.4/.75,3.2/1} {
    \draw[xpenggrid, densely dashed, line width=0.5pt]
      (0,\height) -- (6.1,\height);
    \node[note, anchor=east] at (-0.12,\height) {\ticklabel};
  }
  \draw[xpengblack, line width=0.65pt] (0,0) -- (6.15,0);
  \draw[xpengblack, line width=0.65pt] (0,0) -- (0,3.30);
  \foreach \position/\baseline/\improved/\baselabel/\improvedlabel in {
    0.70/0.93/0.975/.93/.98,
    2.10/0.78/0.98/.78/.98,
    3.50/0.73/0.83/.73/.83,
    5.10/0.8133333333/0.9283333333/.81/.93
  } {
    \fill[xpengbargold] (\position,0) rectangle ++(0.34,{3.2*\baseline});
    \fill[xpengbarblue] (\position+0.40,0) rectangle ++(0.34,{3.2*\improved});
    \node[val, anchor=south] at (\position+0.17,{3.2*\baseline}) {\baselabel};
    \node[val, anchor=south] at (\position+0.57,{3.2*\improved}) {\improvedlabel};
  }
  \foreach \position/\task in {1.07/Banana,2.47/Pour,3.87/Cola,5.47/Avg} {
    \node[note] at (\position,-0.30) {\task};
  }
  \node[font=\fontfamily{phv}\selectfont\scriptsize\bfseries, anchor=south] at (3.075,3.42)
    {(a) Task-progress score};

  \begin{scope}[xshift=7.4cm]
    \fill[xpengavgband, rounded corners=3pt]
      (4.78,-0.50) rectangle (6.08,3.38);
    \foreach \height/\ticklabel in {0/0,0.8/25,1.6/50,2.4/75,3.2/100} {
      \draw[xpenggrid, densely dashed, line width=0.5pt]
        (0,\height) -- (6.1,\height);
      \node[note, anchor=east] at (-0.12,\height) {\ticklabel};
    }
    \draw[xpengblack, line width=0.65pt] (0,0) -- (6.15,0);
    \draw[xpengblack, line width=0.65pt] (0,0) -- (0,3.30);
    \foreach \position/\baseline/\improved/\baselabel/\improvedlabel in {
      0.70/80/90/80/90,
      2.10/50/95/50/95,
      3.50/55/75/55/75,
      5.10/61.66666667/86.66666667/61.7/86.7
    } {
      \fill[xpengbargold] (\position,0) rectangle ++(0.34,{0.032*\baseline});
      \fill[xpengbarblue] (\position+0.40,0) rectangle ++(0.34,{0.032*\improved});
      \node[val, anchor=south] at (\position+0.17,{0.032*\baseline}) {\baselabel};
      \node[val, anchor=south] at (\position+0.57,{0.032*\improved}) {\improvedlabel};
    }
    \foreach \position/\task in {1.07/Banana,2.47/Pour,3.87/Cola,5.47/Avg} {
      \node[note] at (\position,-0.30) {\task};
    }
    \node[font=\fontfamily{phv}\selectfont\scriptsize\bfseries, anchor=south] at (3.075,3.42)
      {(b) Success rate (\%)};
  \end{scope}

  \draw[xpenggrid, fill=white, rounded corners=1pt, line width=0.45pt]
    (4.30,4.02) rectangle (9.45,4.44);
  \fill[xpengbargold] (4.70,4.15) rectangle ++(0.28,0.16);
  \node[note, anchor=west] at (5.10,4.23) {Before DAgger};
  \fill[xpengbarblue] (7.10,4.15) rectangle ++(0.28,0.16);
  \node[note, anchor=west] at (7.50,4.23) {After DAgger};
\end{tikzpicture}}
  \caption{\textbf{Policy self-improvement with world-model-generated recovery data.}
  Before DAgger is the Phase~II-c checkpoint used in the main experiments; After DAgger is its policy-fine-tuned copy.
  We compare banana pick-and-place, water pouring, and cola handover using (a) normalized task-progress score and (b) success rate, with 20 trials per task per method.
  Before DAgger results for banana pick-and-place and water pouring reuse the main-experiment trials, with identical task definitions and scoring.
  The DAgger-fine-tuned policy uses the Phase~II-c data recipe with an $8\%$ mixture of filtered recovery trajectories synthesized by the SGF-adapted Phase~II-c simulator (Table~\ref{tab:dagger_simulator}).
  Shaded columns show unweighted task averages.}
  \label{fig:policy_self_improvement}
\end{figure}

\subsubsection{Policy self-improvement}
\label{sec:policy-self-improvement}
\label{sec:exp-dagger}

Joint world and action modeling enables policy self-improvement: the world
model synthesizes recovery experience that feeds back into policy training.
We evaluate this simulation-to-policy feedback through real-robot performance
after DAgger fine-tuning.  Policy-conditioned simulation further supports
this process by exposing behavioral errors for qualitative diagnosis.

\paragraph{Simulation-driven policy improvement.}
We compare \textbf{Before DAgger}, the Phase~II-c checkpoint used in the main experiments, with \textbf{After DAgger}, a separately fine-tuned policy initialized from that checkpoint.
Only the policy copy is fine-tuned, using a mixture of $8\%$ synthesized recovery examples and $92\%$ data sampled according to the Phase~II-c recipe; the SGF simulator remains fixed.
This fine-tuning stage completes in less than one day.
The SGF-adapted Phase~II-c simulator (Table~\ref{tab:dagger_simulator}) generates deviation--recovery trajectories; consistency-filtered recovery clips are joined with recorded expert tails as supervision, following \S\ref{sec:stage-iii-policy}.
Real-robot evaluation covers banana pick-and-place, water pouring, and cola handover, reporting task-progress scores, success rates, and their unweighted task averages, as shown in Figure~\ref{fig:policy_self_improvement}.

\paragraph{Tasks and evaluation protocol.}
Banana pick-and-place and water pouring use exactly the tasks, scoring rules, and initial conditions defined in \S\ref{sec:exp-ext-ood}; banana success requires placement inside the basket, not grasping alone.
Their Before DAgger results reuse the same evaluation trials reported in Figure~\ref{fig:policy_eval_suite}, rather than a new evaluation.
The comparison follows the main protocol of 20 trials per task per method, with matched initial conditions and resets before and after fine-tuning.
For \textbf{cola handover}, the robot hands the cola to a person standing in front of it.
The task awards one point for each of three cumulative milestones: the hand touches the cola, the cola is lifted, and the cola is handed to the person.
Completing the handover constitutes success and earns the maximum of three points.
We divide each trial's raw score by three and average over the 20 trials to obtain normalized task progress; success rate is the fraction of trials completing all three milestones.

\begin{figure}[!t]
  \centering
  \includegraphics[width=0.9\linewidth]{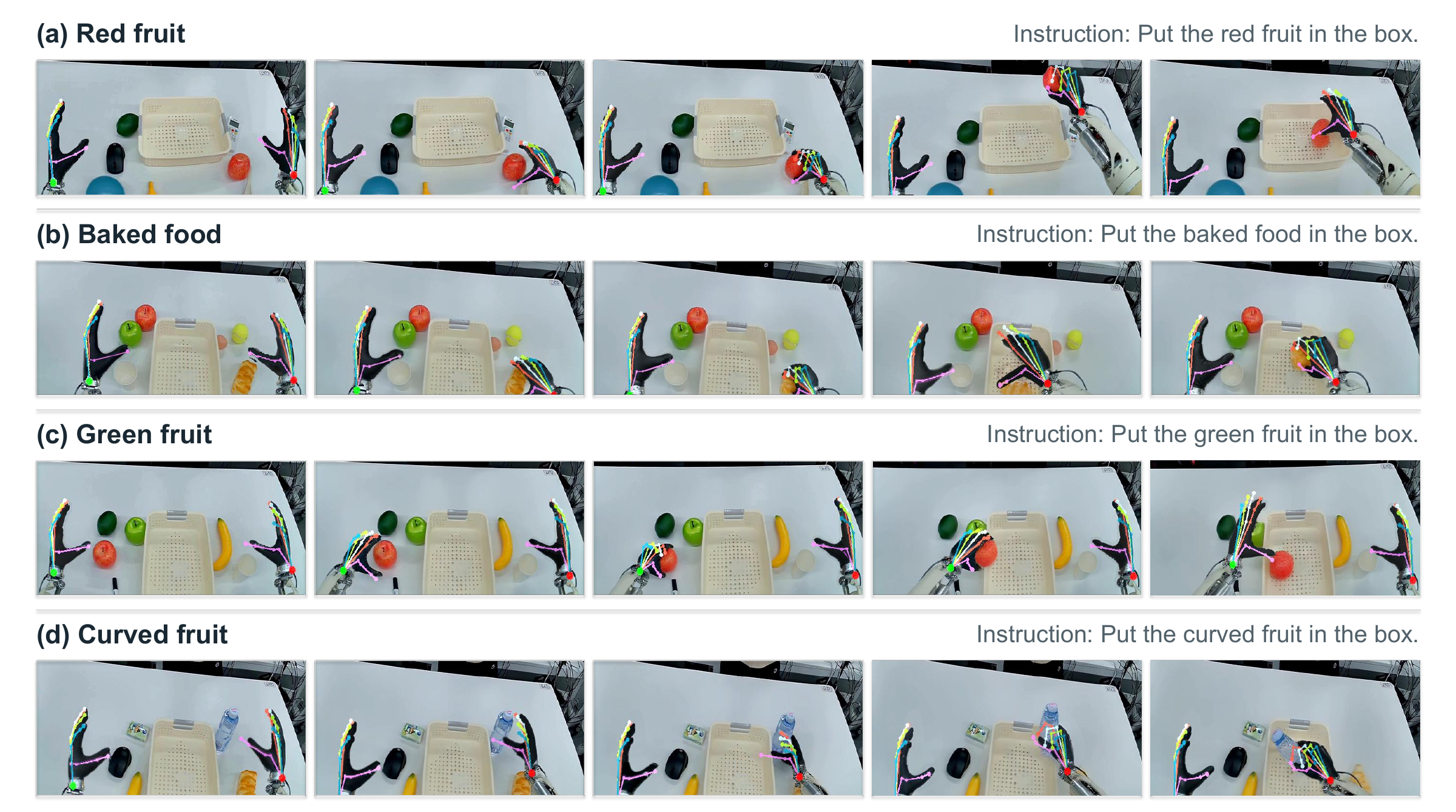}
  \caption{\textbf{Policy--simulation closed-loop evaluation.}
  Policy-predicted actions drive the simulator through skeleton controls and corresponding camera poses, and generated frames are fed back to the policy as subsequent observations.
  Each row shows the instruction and simulated frames with skeleton overlays.
  The first two rows show correct target selection; the last two show a red fruit selected instead of a green fruit, and a bottle instead of a curved fruit.}
  \label{fig:wm_policy_eval}
\end{figure}

\noindent\textbf{Results.}
DAgger fine-tuning consistently improves task-progress scores and success rates across all three tasks.
The mean task-progress score increases from $0.81$ to $0.93$, while the mean success rate increases from $61.7\%$ to $86.7\%$, an absolute improvement of $25$ percentage points.
The largest improvement occurs in water pouring, where task progress
increases from $0.78$ to $0.98$ and success rate from $50\%$ to $95\%$.
Success rates also increase for banana pick-and-place ($80\%$ to $90\%$) and cola handover ($55\%$ to $75\%$), indicating that the benefit is not restricted to a single task.

The concurrent improvements in task progress and success rate indicate
more complete task execution, rather than gains in intermediate milestones
alone.  These real-robot results provide empirical support for
simulation-driven policy self-improvement through fine-tuning with
world-model-generated recovery supervision.
The reported comparison is before versus after fine-tuning; it does not separate recovery-data effects from the additional optimization budget.

\paragraph{Policy--simulation closed-loop diagnosis.}
We inspect policy behavior through closed-loop interaction with the simulator.
Given an observation, the current state, and a language instruction, the policy model predicts actions, which supply skeleton controls and corresponding camera poses for the simulator to generate a predicted video.
The generated frames then serve as observations for the policy model to predict the next actions, forming a repeated policy--simulation closed-loop evaluation.
This procedure supports qualitative assessment of instruction following, target selection, and visible task progress without requiring physical robot execution for each rollout.

Figure~\ref{fig:wm_policy_eval} presents four object-placement examples.
In the first two, the policy selects the requested red fruit and baked food, respectively.
The remaining examples reveal incorrect target selection: a red fruit instead of the requested green fruit, and a bottle instead of the requested curved fruit.
These examples demonstrate that simulated rollouts can identify instruction-following errors even when the generated motions appear plausible.
By reducing the need for robot operation and physical scene resets, simulation provides a lower-cost setting for qualitative policy evaluation and failure analysis before real-robot testing.
\par

\FloatBarrier
\section{Conclusion}
\label{sec:conclusion}

We introduced \modelname, a unified embodied world model that serves as both a world action model and a world simulator through a shared predictive video backbone.
Our results highlight the complementary roles of human experience and robot supervision: human data broaden the interaction knowledge available for learning, while robot demonstrations ground that knowledge in executable control.
Retaining human experience during robot adaptation supports robustness and skill transfer, suggesting that robot demonstrations need not define the full range of behaviors a policy can learn.

Beyond learning from recorded demonstrations, the learned world simulator provides recovery experience for policy improvement and a setting for inspecting policy behavior.
This connects prediction and control not only through shared representations, but also through the use of generated experience to improve subsequent action learning.

Looking ahead, an embodied world model could help a robot decide not only how to act, but also what experience to seek next.
Human observations could suggest unfamiliar skills, imagined rollouts could expose gaps in the policy, and targeted physical interaction could test these predictions and refine both the policy and the simulator.
This would extend the role of simulation from generating recovery examples to guiding exploration, comparing candidate actions, and constructing progressively more demanding learning curricula.
Realizing this vision requires moving beyond the demonstration-centered recovery setting studied here: models must represent uncertainty, remain reliable over longer and contact-rich interactions, and incorporate real-world feedback without forgetting previously acquired skills.
The longer-term opportunity is a robot whose experience is not merely a fixed training resource, but something it can actively expand---learning from people, testing possibilities in imagination, and grounding what it learns in the physical world.

\section*{Contributors}
\label{sec:contributors}

\begingroup
\raggedright
\noindent
    Jiacheng Wei$^{*\S}$,~
    Jerry Bai$^{*\S}$,~
    Xiaoyu Yue$^*$,~
    Zidong Wang$^*$,~
    Xiaoyang Guo$^{*\dagger}$,~
    Cheng Chen,~
    Fanqi Pu,~
    Fan Wu,~
    Zhixu Yue,~
    Yizhuo Li,~
    Feng Qiu,~
    Bo Liu,~
    Yuying Ge,~
    Hui Zhou,~
    Chenyi Chen,~
    Yixiao Ge$^{\ddagger}$
\par\smallskip
  {\noindent
    \footnotesize\color{xpengmuted}%
    $^*$Core contribution.\quad
    $^\S$Equal contribution.\quad
    $^\dagger$Project lead.\quad
    $^\ddagger$Supervision.\quad
    All authors are with XPENG Robotics.\par}

\paragraph{Acknowledgement}
  Yuguo Gan,~ 
  Jianping Li,~  
  Yufan Ren,~
    Kunpeng Song,~
  Wenqian Sun,~ 
    Huimin Pan,~
    Siyang Wang,~
    Xiwen Zhang,~
    Xiaoji Zheng,~
    Ruochong Zheng,~
    Yingji Zhong~

\endgroup

\clearpage
\bibliography{main}
\bibliographystyle{unsrtnat}

\end{document}